%% file: main.tex
\documentclass[a4paper]{article}
\usepackage[a4paper,top=3cm,bottom=2cm,left=3cm,right=3cm,marginparwidth=1.75cm]{geometry}
\usepackage{amsmath}
\usepackage{amssymb}
\usepackage{amsmath}

\usepackage{graphicx}
\usepackage{hyperref}
\usepackage{xurl} 
\usepackage{breakurl}
\usepackage{algorithm,algpseudocode}

\usepackage{color}
\definecolor{CodeGray}{RGB}{240,240,240}
\newcommand{\inlinecode}[1]{\colorbox{CodeGray}{\texttt{#1}}}

\providecommand{\keywords}[1]
{
  \small	
  \textbf{\textbf{Keywords:}} #1
}

\begin{document}

\def\UrlFont{\bfseries}
\title{Optimal To-Do List Gamification for Long Term Planning}
\author{
    Saksham Consul, Jugoslav Stojcheski, Valkyrie Felso, \& Falk Lieder\\
    Rationality Enhancement Group \\
	Max Planck Institute for Intelligent Systems \\
    Tübingen, Baden-Württemberg, Germany
}
\maketitle

\abstract{Most people struggle with prioritizing work. While inexact heuristics have been developed over time, there is still no tractable principled algorithm for deciding which of the many possible tasks one should tackle in any given day, month, week, or year. Additionally, some people suffer from cognitive biases such as the present bias, leading to prioritization of their immediate experience over long-term consequences. The present bias manifests itself as procrastination and inefficient task prioritization. Our method utilizes optimal gamification to help people overcome these problems by incentivizing each task by a number of points that communicates how valuable it is in the long-run. Here, we extend the previous version of our optimal gamification method with additional functionalities for helping people decide which tasks should and should not be done when there is not enough time to do everything. To improve the efficiency and scalability of the to-do list solver, we designed a hierarchical procedure that tackles the problem from the top-level goals to fine-grained tasks. We test the accuracy of the proposed incentivised to-do list by comparing the performance of the strategy with the points computed exactly using Value Iteration for a variety of case studies. These case studies were specifically designed to cover the corner cases to get an accurate judge of performance. Our method yielded the exact same performance as the exact method for all case studies. To demonstrate its functionality, we released an API\footnote{\href{https://github.com/RationalityEnhancement/todolistAPI/tree/multi_smdp_points}{Link to Code}} that makes it easy to deploy our method in Web and app services. We assessed the scalability of our method by applying it to to-do lists with increasingly larger numbers of goals, sub-goals per goal, hierarchically nested levels of subgoals. We found that the method provided through our API is able to tackle fairly large to-do lists with $9$ goals, having a total $\mathbf{576}$ tasks. This indicates that our method is suitable for real-world applications.}

\keywords{optimal gamification; reward shaping; productivity; dynamic programming;
to-do lists; decision-support; hierarchical planning; SMDPs}
\newpage
\tableofcontents
\newpage

\section{Introduction}
Procrastination and prioritization are challenges that many people face in their daily lives \cite{steel2007nature} since setting the right priorities and working diligently requires a lot of mental effort and self-discipline. Procrastination is a consequence of people's tendency choose smaller immediate rewards over larger later rewards \cite{steel2007nature}. This propensity decreases people's productivity, which is defined as the amount of value that a person generates by completing a series of tasks within a fixed amount of time. While several decision support systems have been developed to support human decision-making in specific domains 
\cite{aviv2005partially, bhatnagar1999markov, gadomski2001towards, nunes2009markov,song2000optimal}, only \cite{stojcheski2020optimal} provided a method for helping people decide what to work on on a daily basis. \cite{stojcheski2020optimal}  developed a method to approximate optimal to-do list gamification \cite{lieder2019cognitive} for long to-do lists. Optimal to-do-list gamification incentivizes tasks so as to align the tasks' short-term rewards with their long-term consequences. 

Here, we extend our previous work \cite{stojcheski2020optimal} which laid the foundation to scalable to-do list gamification. \cite{stojcheski2020optimal} exploited the hierarchical structure of to-do list composed of goals and tasks, by using a 2-level hierarchical decomposition of a discrete time semi-Markov decision process (SMDP; \cite{howard1963semi}) and various inductive biases to trim the search space. Moreover, the method mitigated the exponentially-increasing computation cost with respect to the number of tasks in the to-do list. The method incentivizes tasks taking into account the task's deadline, the value the user associates for the completion of the corresponding goal, and the estimated time for completion of the task. The main limitation of the approach was its assumption that all tasks can be completed on time. This led to improper incentivisation when this assumption was violated. Additionally, the previous method was unable to handle to-do lists in which a series of tasks were required to be completed in a specific order. In real life, such cases are common wherein a task requires a specific sequence of actions to be completed. Additionally, people often set unrealistic goals or have goals which are realistic in the beginning but become unrealistic with the passage of time.

In this report, we present an algorithmic solution which extends the work of \cite{stojcheski2020optimal} so as to overcome these limitations and is able to handle more complex planning problems. Our new method supports goal systems with many hierarchically nested levels of subgoals. It allows the user to indicate that some sub-goals are essential to the completion of the corresponding superordinate goals and assign different levels of importance to different goals and tasks. Furthermore, it can handle tasks that are intrinsically valuable to the user regardless of their contribution to achieving a larger goal. To compute the incentives for such to-do list, our new method breaks down the to-do list into smaller SMDPs (called mini-SMDPs) comprised of goals and sub-goals, which are then solved using the 2-level hierarchical abstraction for the mini-SMDP. The same method is applied recursively, as the values of the sub-goals are propagated downwards. The main benefit of the added parameters is that the assignment of optimal points take into account more human-like considerations and are more in line with a user's intentions. It also allows for dependencies between tasks imposed by users in order to form sequences
of tasks that have to be executed in a particular order by creating a chain of sub-goals, with higher-priority tasks kept lower in the chain,  all marked essential for completing its immediate superior sub-goal.  

Furthermore, we packaged this algorithm in an application programming interface (API) that takes in a to-do list and outputs a gamified list of tasks.

This text is divided into the following sections: In Section~\ref{sec:formulation}, we describe the formal definition of the problem as a discrete-time SMDP. In Section~\ref{sec:method}, the algorithm is explained with Section~\ref{sec:miniSMDP} describing how the mini-SMDP are solved and Section~\ref{sec:value} describes how the value of the higher level mini-SMDP is transferred to the lower-level mini-SMDP. In Section~\ref{sec:results}, we
provide a case study to show the the assignment of pseudo rewards and illustrate its functionality by  comparing the performance of our proposed algorithm with an exact solution using Value Iteration \cite{bellman1957markovian}. We systematically show how our proposed algorithm has no difference in performance as the exact solution. Additionally, Section~\ref{sec:results} also evaluates the speed and reliability of the
API. Section~\ref{sec:API} introduces the API we have developed. Lastly, in Section~\ref{sec:future} we outline directions for future work.

\section{Problem Formulation}\label{sec:formulation}
The goal of to-do list gamification is to maximize the user’s long-term productivity by proposing
daily task lists where each task is incentivized by a certain number of points. These points transform the short term rewards to reflect the long term rewards and consequences overcoming human's present bias \cite{steel2007nature}. Users compose
hierarchical to-do lists comprised of three types of items: \emph{goals}, \emph{sub-goals} and \emph{tasks}. We define a \textit{task} as a sub-goal that cannot be further decomposed. Simply put, a task is the smallest non-divisible unit of work.

\begin{figure}[!t]
        \centering
        \includegraphics[width=0.95\textwidth]{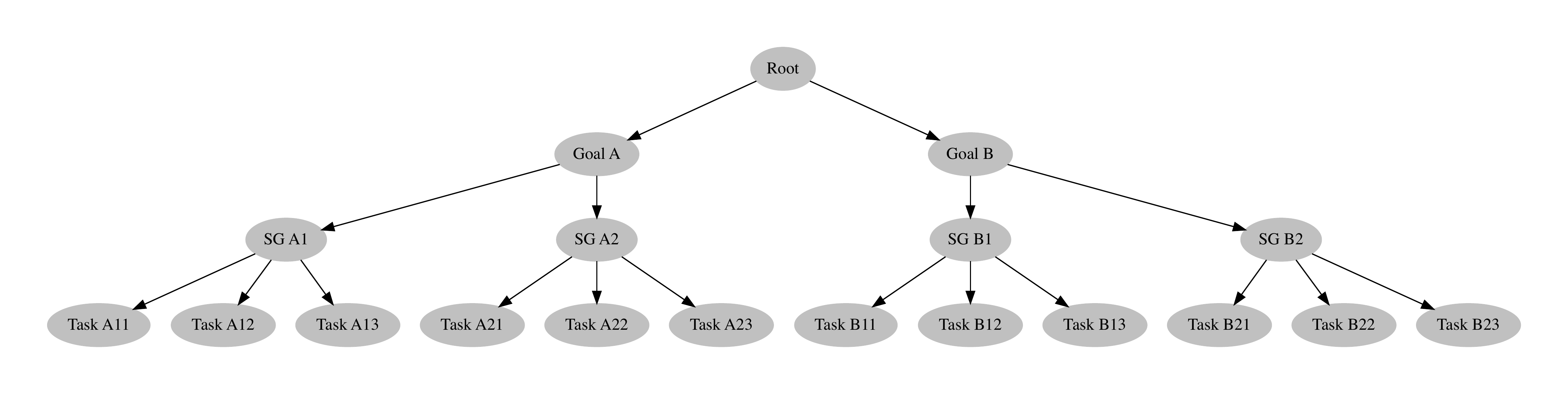}
        \caption{Graphical example of a hierarchical to-do-list.}
        \label{fig:graph}
    \end{figure}

Solving a SMDP generates the optimal plan to complete a list of user-specified tasks. The solution of the large SMDP is approximately computed by solving each level of the hierarchical to-do list as a mini-SMDP. Each mini-SMDP consists of $1$ \emph{goal} and multiple \emph{sub-goals}. In each mini-SMDP, the sub-goal is treated as a task, and as such, would be referred as a task from now on. We define a root of a to-do list as an imaginary node, whose sub-goals are  top-most goals as seen in Figure~\ref{fig:graph}. \textbf{Goals} consist of a deadline, a value estimate, and a list of sub-goals. Each \textbf{sub-goal} contains a time estimate to complete the sub-goal, a Boolean value to indicate if the sub-goal is essential to complete its corresponding goal, its importance to complete its immediate goal, and intrinsic reward, which is the reward obtained on completing a sub-goal, independent to the completion of its goal. A sub-goal is deemed as essential if the completion of the sub-goal is necessary for the completion of the goal. In other words, a goal cannot be completed without completing \textit{all} of its essential sub-goals. As such, an essential sub-goal is marked with a high importance factor, and a non-essential sub-goal should be marked with a low importance factor.

In addition, users specify their desired workload in hours for a typical day (\emph{typical day's working hours}) and for the day at hand (\emph{today's working hours}). The outputted gamified daily schedule should contain all tasks that users indicated they wanted to work on today, as well as additional tasks towards their goals, up to the desired daily workload.

\subsection{Modelling the to-do-list as a discrete-time SMDP}
The state space $\mathcal{S}$ consists of all possible combinations of completed and uncompleted tasks. Since a task can either be completed or uncompleted, each $s \in \mathcal{S}$ is a binary vector of length $n+1$, where $n$ corresponds to the total number of tasks in the SMDP and the slack-off task. The slack-off task represents when an user decides to no longer do any more productive work. If the $i$-th element of any binary vector $s \in \mathcal{S}$ is $1$, the task associated with the element is completed, and $0$ otherwise. Similarly, the action space $\mathcal{A}$ refers to the selection of a task and hence $a \in \mathcal{A}$ is of the same size, $n+1$.

\begin{equation}
    \mathcal{S} = \{\{ 0, 1\}^{n+1}\}
\end{equation}

Considering $G$ top-most goals, each goal having $D$ sub-goal levels and each sub-goal having $B$ sub-sub goals, the to-do list would have a total of  $G \cdot B^D$ tasks. Without breaking the to-do list into multiple mini-SMDPs, the state space would be of size $2^{(G \cdot B^D+1)}$. By breaking down into mini-SMDPs, we would need to solve $G \cdot D$ SMDPs with a state space $2^{B+1}$. Hence, the decomposition of the to-do list into mini-SMDPs shrink the size of the state space and the number of computations go down from $\mathbf{2^{(G \cdot B^D+1)}}$ to  $\mathbf{B\cdot D \cdot 2^{B+1}}$.

\subsubsection{Transition time $F$}  \label{sec:smdp_time}
In a SMDP setting, the transition-time function,  $F(\tau | s_{t}, a)$ is the probability that the time at which the agent has to make the next decision occurs in exactly $\tau$ time units, as a consequence of executing action $a$ in state $s$ at time $t$.
We chose a transition-time function that can model the cognitive bias known as the \textit{planning fallacy}, in which people underestimate the time required to complete a task. Kahneman and Tversky \cite{kahneman1977intuitive} describe this bias as such: "\textit{Scientists and writers, for example, are notoriously prone to underestimate the time required to complete a project, even when they have considerable experience of past failures to live up to planned schedules... It frequently occurs even when underestimation of duration or cost is actually penalized}." 

Since the SMDPs are modelled with discrete time steps and people have unreliable time estimates, we model the number of time units required for action completion to follow a zero-truncated Poisson probability distribution\footnote{Also known as \textit{conditional Poisson distribution}, \textit{positive Poisson distribution}.} with adjusted mean value and variance. We formally define the transition-time function as
\begin{equation*}
    F(\tau | s_{t}, a)
    := \text{Poisson}_{> 0}(\tau; \tilde{k} )
    = \dfrac{ \tilde{k}^{\tau} e^{- \tilde{k} } }{ \tau! (1 - e^{- \tilde{k} }) }
\end{equation*}
where $\tilde{k} = c_{\text{pf}} \cdot k$, $k$ is the discrete amount of time units required to complete action $a$ in state $s$ at time $t$, and $c_{\text{pf}} \in \mathbb{R}_{> 0}$ is a planning-fallacy constant that adjusts the distribution parameter. In lack of knowledge about the exact value of the planning-fallacy constant ($c_{\text{pf}}$), we follow King and Wilson \cite{king1967subjective} and we initially set its value to $1.39$. Obtaining better estimates for this value based on real-world data is left for future work.

\subsubsection{Transition dynamics $T$}  \label{sec:smdp_transitions}
The transition dynamics from a current state $s$ at time $t$ to a next state $s'$ after executing an action $a$ is deterministic in completion, but stochastic in duration. In other words, the presented algorithm assumes that users will complete a task once they they start but may require more time than the time estimated for completion of the task.
Under these assumptions,  In other words, the transition probability $T(s_{t}, a, s_{t + \tau}')$ is completely dependent on the probability of completing an action in exactly $\tau$ time units, which can be formally written as
\begin{equation*}
    T(s_{t}, a, s_{t + \tau}')
    = \text{Pr}(s_{t+\tau}' | s_{t}, a) \sim F(\tau | s_{t}, a)
    \hspace{0.4cm} \forall s \in \mathcal{S}_{\text{(task)}}^{(g)}
    \hspace{0.4cm} \forall a \in \Omega_{\text{(task)}}^{(g)}
    \hspace{0.4cm} t \in \mathbb{Z}_{\ge 0}
    \hspace{0.4cm} g = 1, \ldots, |\mathbf{\mathcal{G}}|
\end{equation*}
where $\tau \sim F(\tau | s_{t}, a)$ is the transition-time function that determines the time needed to complete a chosen action $a$ in state $s$ at time $t$, and $s'$ is \textit{the} state that follows as a consequence of executing action $a$ in state $s$. Formally, if an action $a$ is represented by the $i$-th bit of the binary state vector $s$, the binary vector of the next state $s'$ can be written as 
$s' = e_{i} \vee s$,
where $e_{i}$ is a one-hot vector with a value of $1$  only at its $i$-th position, and $\vee$ represents the ``or'' operation of two binary vectors. 

A special case of the transition dynamics occurs after reaching the terminal state in which all real tasks have been completed or if the slack-off action is selected. There, the process transits to a goal-achieving state $s_{\dagger}$ after instantaneously executing the action  $a_{\dagger}$  in $0$ time steps, that is
$
    T(\mathbf{1}, a_{\dagger}, s_{\dagger})
    = \text{Pr}(s_{\dagger} | \mathbf{1}, a_{\dagger})
    = 1
$.
\subsubsection{Reward function}
We define the reward function $r(s_t, a, s'_{t+\tau})$ from a current state $s$ at time $t$ to the next state $s'$ at time $t+\tau$ after performing action $a \in \mathcal{A}$ that takes $\tau$ time units for execution to be
\begin{multline}
    r(s_t, a, s'_{t+\tau}) = 
    \begin{cases}
    R(a_+) \cdot (1-\gamma)^{-1} &\mbox{if the slack off action was chosen}\\
    -\lambda^{(g)}\sum_{k=0}^{\tau-1}\gamma^k \\+ \gamma^{\tau-1} \cdot r_{\textrm{extrinsic}}(s_{t}, a, s'_{t + \tau}) \cdot \Pi(\beta_{g}) \\+ r_{\mathrm{intrinsic}}( a) &\mbox{if any other action was chosen}
    \end{cases}
\end{multline}

\begin{equation}
    r_{\mathrm{(extrinsic)}}^{(g)}(s_{t}, a, s'_{t + 
    \tau}) = 
    \begin{cases}
    R(g) \cdot \frac{\sum{I_{done}(s'_{t + \tau})}}{\sum_{a_k \in \mathbb{A}^{g}}I(a_k)}  &\mbox{if goal is complete}\\
    0 &\mbox{if goal hasn't been completed}\\
    \end{cases}
\end{equation}

where $a_+$, represents \emph{slack-off} action, $\gamma \in (0,1]$ is a discount factor, $R(g)$ indicates the value of a goal $g \in \mathcal{G}$. $\lambda^{(g)} \in \mathbb{R}_{>0}$ models the value that reflects the cost of a person’s time and mental effort to work on goal $g$. $I_{done}(s'_{t + \tau})$ refers to the list of importance values of the subset of completed tasks in state $s'_{t + \tau}$. 

$R(g)$ returns the goal value if executing the next task-level action $a$ leads to completion of goal $g$. We define $\Pi(\beta_{g})$ to be the penalty function for a goal $g$. The value
of the penalty function discounts the goal reward proportionally to the time by which deadlines
associated with that goal are missed and it can be formally written as $\Pi(\beta_{g}) = (1+\beta_g)^{-1}$. Here $\beta_g = \sum_{i}^{n}\psi \cdot \Delta t_i$ is a weighted sum of penalties for tasks whose deadlines were $\psi \in \mathbb{R}_{>0}$ is the penalty rate (per unit time) and $\Delta t_i$ is the number of time units by which the deadline was missed.

According to the definition, an immediate negative reward and a small positive reward is obtained for completing each task. Conversely, an immediate positive reward is obtained \emph{after} the goal is completed or a slack-off action has been chosen for execution. A goal is said to be completed if all \emph{essential} sub-goals are completed. 

\subsubsection{Optimal policy}
As described in Section~\ref{sec:value}, each to-do list is broken down into mini-SMDPs in a hierarchical manner, treating each layer as a goal to propagate and compute the value of each individual task. Once all of the 2-layer mini-SMDPs have been solved, their tasks and the corresponding point values are collated into a single gamified to-do list. The (approximately) optimal policy can then be defined as always choosing the task with largest number of points from the list of uncompleted tasks.

\section{Method}\label{sec:method}
An SMDP consists of one goal and multiple sub-goals. As mentioned in Section~\ref{sec:formulation}, while solving the SMDP, we consider each sub-goal as a task with no sub-tasks.

\subsection{Solving mini-SMDP}\label{sec:miniSMDP}
Given a mini-SMDP with $B$ sub-tasks and starting time as $t = t_o$, the stating state $s_{t_o}$ is represented as a vector of size $B+1$. The mini-SMDP is computed by first checking the tasks completed, marking the vector $1$ if the corresponding task is completed. The last index of the vector represents if the slack-off action has been previously selected, indicating that the state is now in state $s_{\dagger}$.

From the starting state, the method computes the expected reward for all the possible sequences of tasks. This is implemented using the wrapper function \textsc{solve}(Algorithm~\ref{alg:solve}) and recursively calling \textsc{solveNext}(Algorithm~\ref{alg:solvenext}) which utilizes \textsc{getExpectedReward}(Algorithm~\ref{alg:expReward}). The method \textsc{solveNext}(Algorithm~\ref{alg:solvenext}) shows the recursive calls to all possible sequences of tasks. After initializing the $Q$-values for a given state and time, it iterates over all the possible actions in the given state. If the optimal sequence following an action is not computed before and if the action is not a slack-off action, the method computes the expected reward for following the given action and then the optimal policy before updating the $Q$-value for a given state and the expected reward of doing a task.

The method \textsc{getExpectedReward}(Algorithm~\ref{alg:expReward}) finds the expected reward for a given action. It shows how the SMDP is solved, taking into account that each task is completed in an unequal amount of time. The expected reward for completing the task is computed by iterating over time estimates using the transition function $F$. The expected task reward is calculated by taking the weighted average of the reward obtained by completing a task for each time estimate. It takes into account the penalty of missing the deadline (if missed) and the discounted cumulative cost for performing the task an the immediate reward from the reward function $r(s_t, a, s'_{t'})$. Additionally, the reward for following the optimal policy is computed for each time estimate to also compute the expected total reward used  Used for updating the $Q$-value.

\begin{algorithm}[p]
 \caption{Solving the mini-SMDP}\label{alg:solve} 
\begin{algorithmic}
\Require{$s_{t_o}$, $t_{o}$} 
\Ensure{$\mathrm{P}, \mathrm{Q}$} \Comment{Returning the expected reward \& policy from the given state}
\Procedure{solve}{$s_{t_o}, t_o$} 
 \State $Q = \text{\textsc{solveNext}}(s_{t_o}, t_o)$
 \State $a, _ = max(Q[s_{t_o}][t_o])$ \Comment{The optimal action to do in current state}
 \State $\mathrm{P}[s_{t_o}][t_o] = a$ 
 \EndProcedure
 \end{algorithmic}
\end{algorithm}

\begin{algorithm}[p]
 \caption{Recursive function to compute expected reward for given state and time.} \label{alg:solvenext}
\begin{algorithmic}
\Require{$s_{t}$, $t$} 
\Ensure{$\mathrm{P}, \mathrm{Q}$} \Comment{Returning the expected reward and policy from the given state}
\Procedure{solveNext}{$s_{t}, t$} 
 \State $\text{Initialize dictionaries for Q-function for given $s_{t}, t$}$
 \For{$a \gets \textsc{nextPossibleAction}(s_{t})$}
 \State $s'_{t} = \textsc{nextState}(s_{t}, a)$
 \If{$a \in Q[s_{t}][t].\mathrm{keys}$}\Comment{Computation already computed}
\State $\mathrm{continue}$ 
\EndIf
\If{$a \neq \text{slack-off action}$}
\State $t_{\mathrm{max}} = \textsc{getDeadline}(a)$
\Comment{$t_{\mathrm{max}}$ is the time remaining to deadline}
\State $ \mathrm{expTaskReward}, \mathrm{expTotalReward} = \textsc{getExpectedReward}(s_{t}, t, a, t_{\mathrm{max}})$
\State $Q[s_{t}][t][a] \mathrel{{+}{=}} \mathrm{expTotalReward}$
\State $R[s_{t}][t][a] \mathrel{{+}{=}} \mathrm{expTaskReward}$
\EndIf
 \EndFor
 \EndProcedure
 \end{algorithmic}
\end{algorithm}

\begin{algorithm}[p]
\caption{Computing the expected reward of doing action $a$ in state $s$ in time $t$} \label{alg:expReward}
\begin{algorithmic}
\Require{$a, s_{t}, t, t_{\mathrm{max}}$} 
\Ensure{$\mathrm{expTaskReward}, \mathrm{expTotalReward}$}
\Procedure{getExpectedReward}{$(s_{t}, t, a, t_{\mathrm{max}}$} 
 \State $\beta = 0$
 \State $\mathrm{expTotalReward} = 0$
 \State $\mathrm{expTaskReward} = 0$
 \For{$\tau,  F(\tau|s_{t}, a) \gets  
 \textsc{getTimeTransitions}(s_{t}, a)$}
  \State $t' = t + \tau$
 \State $ r = \lambda_{g} \cdot \sum_{i=1}^{\tau}\gamma^{i-1}$  \Comment{Computing total loss of performing action}
  \If{$\textsc{deadlineMissed}(t')$}
  \State $\beta \mathrel{{+}{=}} F(\tau|s_{t}, a) \cdot [\psi \cdot (t' - t_{\mathrm{max}})]$
  \EndIf
  \State $r \mathrel{{+}{=}} r(s_t, a, s'_{t'}) $
  \State $\mathrm{expTaskReward} \mathrel{{+}{=}} F(\tau|s_{t}, a) \cdot r$
  \State $\mathrm{Q}_{\_}$ = $\textsc{solveNext}(s'_{t'}, t')$
  \State $a', r' = max(Q_{\_}[s'_{t'}][t'])$
 \State $\mathrm{P}[s'_{t'}][t'] = a'$ 
 
 \State $\mathrm{expTotalReward} \mathrel{{+}{=}}F(\tau|s_{t}, a) \cdot \left(r + \gamma^{\tau} \cdot r' \right)$ 
  \EndFor
  \EndProcedure
 \end{algorithmic}
\end{algorithm}

\subsection{Passing of the value function}\label{sec:value}
To understand how the value is propagated down the SMDP, we utilize an example to-do list shown in Figure~\ref{fig:graph}. It consists of $2$ goals, which having $2$ sub-goals. Each sub-goal has $3$ tasks. The SMDP is broken down into $7$ mini-SMDPs as shown below: 
\begin{enumerate}\label{list}
    \item Goal: Root, Sub-goals: Goal A, Goal B
    \item Goal: Goal A, Sub-goals: SG A1, SG A1
    \item Goal: Goal B, Sub-goals: SG B1, SG B2
    \item Goal: SG A1, Sub-goals: Task A11, A12, A13
    \item Goal: SG A2, Sub-goals: Task A21, A22, A23
    \item Goal: SG B1, Sub-goals: Task B11, B12, B13
    \item Goal: SG B2, Sub-goals: Task B21, B22, B23
\end{enumerate}
A dummy "Root" node is created to facilitate solving of the mini-SMDP. The root goal value is the sum of values of the goals assigned by the user and denotes the total number of productivity value in the to-do list. The importance of each goal is given as the ratio between the value assigned to the goal and the sum of values of all goals. 

In the $1^{\mathrm{st}}$ mini-SMDP, all goals are represented as tasks and marked as non-essential. The intrinsic reward of a goal is computed recursively by taking the sum of the intrinsic reward of all its sub-goals.

After solving the $1^{\mathrm{st}}$ mini-SMDP, the value of the sub-goals (Goal A, Goal B), denoted by $R(SG_k)$ will be passed down to the $2^{\mathrm{nd}}$ and $3^{\mathrm{rd}}$ mini-SMDPs as follows
\begin{equation}
    R(SG_k) = \frac{e^{\eta(SG_k)}}{\sum_{i=1}^{n}e^{\eta(SG_i)}} \cdot \left[R(g) + \sum_{\forall SG_i\in P(g) }r_{\textrm{intrinsic}}(SG_i) \right]
\end{equation}
with
\begin{equation}
    \eta(SG_k) =  \{\gamma^{\tau_{SG_k}} \cdot \mathbb{E}[V^{*}(s'|s,SG_k)] - V^{*}(s)\} + r_{\textrm{extrinsic}}(s_{t}, SG_k, s'_{t + \tau_{SG_k}}) \cdot \Pi(\beta_{SG_k}) + r_{\mathrm{intrinsic}}(s_{t}, SG_k, s'_{t + \tau_{SG_k}})
\end{equation}
where $R(g)$ is the return of the optimal policy of the $1^{st}$ mini-SMDP. $\tau_{SG_k}$ is the time estimate to complete all essential tasks of the sub-goal $SG_k$.

\begin{figure}[!t]
        \centering
        \includegraphics[width=0.95\textwidth]{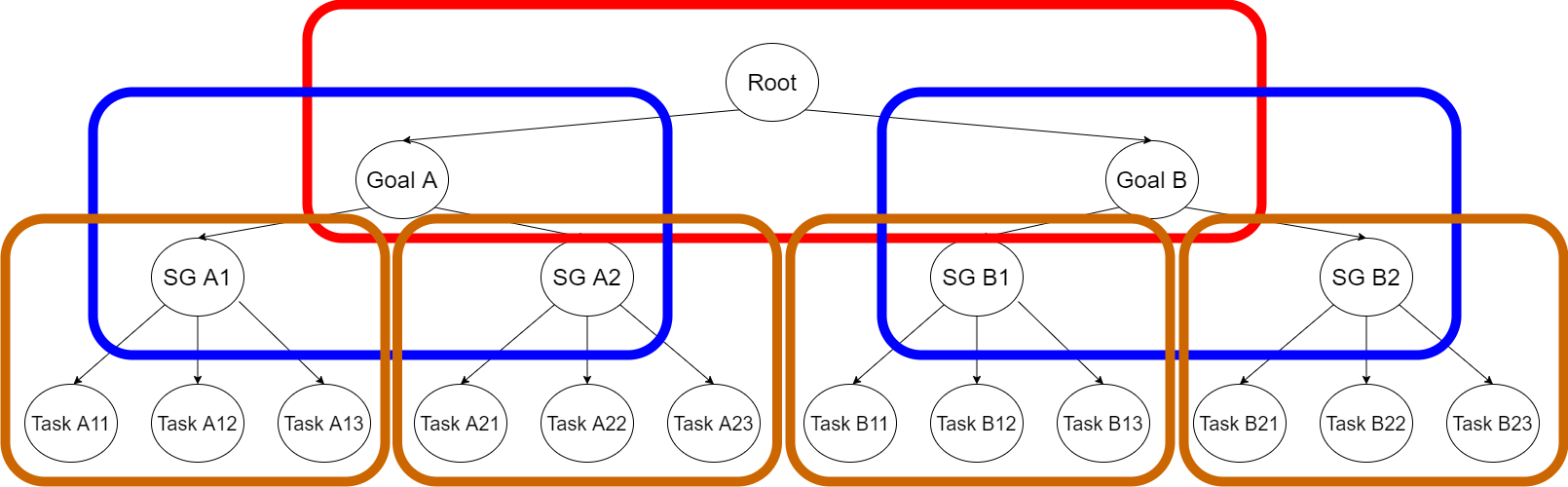}
        \caption{Graphical depiction of how the mini-SMDPs are solved. The mini-SMDP in red is solved first, followed by the ones in blue and finally the mini-SMDPs in brown are solved last.}
        \label{fig:solve}
    \end{figure}

Similarly, after solving the $2^{\mathrm{nd}}$ mini-SMDP, the value of SG A1 will be passed down to the $4^{\mathrm{th}}$ mini-SMDP and value of SG A2 to $5^{\mathrm{th}}$ mini-SMDP. Likewise, the value of SG B1 and SG B2 will be passed down to the $6^{\mathrm{th}}$ and $7^{\mathrm{th}}$ mini-SMDP, respectively. The flow of solving mini-SMDPs is depicted in Figure~\ref{fig:solve}

\section{Results}\label{sec:results}
In order to ensure the accuracy of the computed optimal rewards, we first generate a case study and discuss the algorithm's output. Specifically, in Section~\ref{sec:specific-eg}, we present the optimal sequence of tasks suggested by our proposed algorithm. 

Furthermore, in Section~\ref{sec:compare}, we compare the performance of the proposed method with an exact method to compute the optimal points for a number of case studies.
In Section~\ref{sec:scale}, we show the limits of the proposed algorithm which satisfies the $30$ seconds time limit imposed by the Heroku service.

\subsection{A case study of optimal to-do list gamification}\label{sec:specific-eg}
In the following test case, we designed a realistic example with 3 goals. Goal A is the goal with the closest deadline, which is expected to be completed only if all its tasks are prioritized first. Goal B, is the next goal which has a more flexible deadline and can be completed even after Goal A has been completed. Goal C is a goal which has a high value but it is impossible to be completed within its deadline. The detailed description of the to-do list is given below. For each goal, a \texttt{Value} is assigned which represents the value associated for completion of a goal, similarly, there is a \texttt{Deadline} which indicates how many time-units the user has to complete the goal. Each sub-goal has an \texttt{Intrinsic Reward} which denotes the reward derived of completing the sub-goal, independent of the corresponding goal. \texttt{Ess} is a Boolean value which is \texttt{True} is the sub-goal is deemed essential to complete the corresponding goal. Additionally, 
\texttt{Imp} denotes how important the sub-goal is relative to the other sub-goals. Finally, \texttt{Time} indicates the estimated time to complete a sub-goal.
\begin{verbatim}
    
Goal A, - Value: 1000, Deadline: 12,

     SG A1 - Intrinsic Reward: 40, Ess: True,Imp: 100, Time: 7

       Task A11 - Intrinsic Reward: 10, Ess: True,Imp: 60, Time: 3

       Task A12 - Intrinsic Reward: 15, Ess: True,Imp: 20, Time: 2

       Task A13 - Intrinsic Reward: 15, Ess: True,Imp: 20, Time: 2

     SG A2 - Intrinsic Reward: 30, Ess: True,Imp: 100, Time: 5

       Task A21 - Intrinsic Reward: 20, Ess: True,Imp: 60, Time: 3

       Task A22 - Intrinsic Reward: 2, Ess: True,Imp: 30, Time: 1
         
       Task A23 - Intrinsic Reward: 8, Ess: True,Imp: 10, Time: 1
         
Goal B, - Value: 500, Deadline: 50,

     SG B1 - Intrinsic Reward: 100, Ess: True,Imp: 100, Time: 6

       Task B11 - Intrinsic Reward: 80, Ess: True,Imp: 90, Time: 4

       Task B12 - Intrinsic Reward: 20, Ess: True,Imp: 10, Time: 2

     SG B2 - Intrinsic Reward: 100, Ess: True,Imp: 100, Time: 17

       Task B21 - Intrinsic Reward: 20, Ess: True,Imp: 20, Time: 2

       Task B22 - Intrinsic Reward: 10, Ess: True,Imp: 60, Time: 2

       Task B23 - Intrinsic Reward: 10, Ess: True,Imp: 2, Time: 1
         
       Task B24 - Intrinsic Reward: 40, Ess: True,Imp: 15, Time: 10

       Task B25 - Intrinsic Reward: 20, Ess: True,Imp: 3, Time: 2
         
Goal C, - Value: 5000, Deadline: 50,

     SG C1 - Intrinsic Reward: 10, Ess: True,Imp: 100, Time: 3

       Task C11 - Intrinsic Reward: 5, Ess: True,Imp: 60, Time: 1
         
       Task C12 - Intrinsic Reward: 5, Ess: True,Imp: 40, Time: 2
         
     SG C2 - Intrinsic Reward: 90, Ess: True,Imp: 100, Time: 502

       Task C21 - Intrinsic Reward: 50, Ess: True,Imp: 20, Time: 50

       Task C22 - Intrinsic Reward: 10, Ess: True,Imp: 60, Time: 400

       Task C23 - Intrinsic Reward: 20, Ess: True,Imp: 10, Time: 50

       Task C24 - Intrinsic Reward: 10, Ess: True,Imp: 10, Time: 2
\end{verbatim}
The optimal sequence of tasks is found by selecting tasks in a myopic greedy manner, in which the task with the highest points is selected first and is completed exactly in the estimated time. The optimal sequence of tasks in the to-do list is to first select all the tasks to achieve Goal A. Since all tasks of Goal A are essential, all tasks of Goal A are first selected. Afterwhich, the tasks to achieve Goal B were selected. Following which, only the tasks needed to accomplish Goal C remain. At this stage, the optimal action to perform would be to slack-off. The slack-off reward is $10.11$ which has more points than the any task in Goal C, with the highest points belonging to Task C$11$, with $7.0$ points. The selection to slack-off makes sense and can be explained by looking into the sub-goals of Goal C. Goal C consists of $2$ essential sub-goals, of which sub-goal SG C$2$, is not possible to be completed realistically before the deadline, hence, performing any tasks of Goal C would not yield to the completion of Goal C and should not be completed. 

The printout below shows the gamified to-do list that would be shown to the user for the specifc case study, where the user selects the task with the highest points and performs it in its estimated time. In each printout, \texttt{Tasks Completed:}, lists the completed tasks.
Following which, the list of uncompleted tasks are listed in descending order of points, where \texttt{PRS} denote the amount of points computed for the corresponding task. Finally, \texttt{Net PR Sum:} gives the sum of points, rounded to the nearest integer, of all uncompleted tasks.
\begin{itemize}
    
\item{
\begin{verbatim}Tasks Completed:  []
Task: Task A12, PRS: 1574.467
Task: Task A13, PRS: 1574.467
Task: Task A11, PRS: 1574.308
Task: Task A23, PRS: 322.849
Task: Task A22, PRS: 322.848
Task: Task A21, PRS: 322.79
Task: Task B25, PRS: 282.829
Task: Task B21, PRS: 282.829
Task: Task B23, PRS: 282.829
Task: Task B22, PRS: 282.824
Task: Task B24, PRS: 282.637
Task: Task B11, PRS: 93.997
Task: Task B12, PRS: 93.989
Task: Task C11, PRS: 7.0
Task: Task C12, PRS: 6.999
Task: Task C24, PRS: -399.847
Task: Task C21, PRS: -399.887
Task: Task C23, PRS: -400.043
Task: Task C22, PRS: -402.878
Net PR Sum: 5705
\end{verbatim}}
\item{
\begin{verbatim}Tasks Completed:  ['Task A12']
Task: Task A13, PRS: 1616.077
Task: Task A11, PRS: 1615.915
Task: Task B25, PRS: 282.829
Task: Task B21, PRS: 282.829
Task: Task B23, PRS: 282.829
Task: Task B22, PRS: 282.824
Task: Task B24, PRS: 282.637
Task: Task A23, PRS: 268.55
Task: Task A22, PRS: 268.55
Task: Task A21, PRS: 268.503
Task: Task B11, PRS: 93.997
Task: Task B12, PRS: 93.989
Task: Task C11, PRS: 7.0
Task: Task C12, PRS: 6.999
Task: Task C24, PRS: -399.847
Task: Task C21, PRS: -399.887
Task: Task C23, PRS: -400.043
Task: Task C22, PRS: -402.878
Net PR Sum: 4051
\end{verbatim}}
\item{
\begin{verbatim}
Tasks Completed:  ['Task A12', 'Task A13']
Task: Task A11, PRS: 1613.196 
Task: Task B25, PRS: 282.829 
Task: Task B21, PRS: 282.829 
Task: Task B23, PRS: 282.829 
Task: Task B22, PRS: 282.824 
Task: Task B24, PRS: 282.637 
Task: Task A23, PRS: 258.754 
Task: Task A22, PRS: 258.753 
Task: Task A21, PRS: 258.708 
Task: Task B11, PRS: 93.997 
Task: Task B12, PRS: 93.989 
Task: Task C11, PRS: 7.0 
Task: Task C12, PRS: 6.999 
Task: Task C24, PRS: -399.847 
Task: Task C21, PRS: -399.887 
Task: Task C23, PRS: -400.043 
Task: Task C22, PRS: -402.878 
Net PR Sum: 2403
    
\end{verbatim}}
\item{
\begin{verbatim}
Tasks Completed:  ['Task A12', 'Task A13', 'Task A11']
Task: Task A23, PRS: 939.814 
Task: Task A22, PRS: 939.813 
Task: Task A21, PRS: 939.632 
Task: Task B21, PRS: 282.829 
Task: Task B25, PRS: 282.829 
Task: Task B23, PRS: 282.829 
Task: Task B22, PRS: 282.824 
Task: Task B24, PRS: 282.637 
Task: Task B11, PRS: 93.997 
Task: Task B12, PRS: 93.989 
Task: Task C11, PRS: 7.0 
Task: Task C12, PRS: 6.999 
Task: Task C24, PRS: -399.847 
Task: Task C21, PRS: -399.887 
Task: Task C23, PRS: -400.043 
Task: Task C22, PRS: -402.878 
Net PR Sum: 2833
\end{verbatim}}
\item{
\begin{verbatim}
Tasks Completed:  ['Task A12', 'Task A13', 'Task A11', 'Task A23']
Task: Task A22, PRS: 932.907 
Task: Task A21, PRS: 932.726 
Task: Task B21, PRS: 282.829 
Task: Task B25, PRS: 282.829 
Task: Task B23, PRS: 282.829 
Task: Task B22, PRS: 282.824 
Task: Task B24, PRS: 282.637 
Task: Task B11, PRS: 93.997 
Task: Task B12, PRS: 93.989 
Task: Task C11, PRS: 7.0 
Task: Task C12, PRS: 6.999 
Task: Task C24, PRS: -399.847 
Task: Task C21, PRS: -399.887 
Task: Task C23, PRS: -400.043 
Task: Task C22, PRS: -402.878 
Net PR Sum: 1879
\end{verbatim}}
\item{
\begin{verbatim}
Tasks Completed:  ['Task A12', 'Task A13', 'Task A11', 'Task A23', 'Task A22']
Task: Task A21, PRS: 932.0 
Task: Task B21, PRS: 282.829 
Task: Task B25, PRS: 282.829 
Task: Task B23, PRS: 282.829 
Task: Task B22, PRS: 282.824 
Task: Task B24, PRS: 282.637 
Task: Task B11, PRS: 93.997 
Task: Task B12, PRS: 93.989 
Task: Task C11, PRS: 7.0 
Task: Task C12, PRS: 6.999 
Task: Task C24, PRS: -399.847 
Task: Task C21, PRS: -399.887 
Task: Task C23, PRS: -400.043 
Task: Task C22, PRS: -402.878 
Net PR Sum: 945
\end{verbatim}}
\item{
\begin{verbatim}
Tasks Completed:  ['Task A12', 'Task A13', 'Task A11', 'Task A23', 'Task A22', 'Task A21']
Task: Task B21, PRS: 982.306 
Task: Task B25, PRS: 982.306 
Task: Task B23, PRS: 982.306 
Task: Task B22, PRS: 982.301 
Task: Task B24, PRS: 981.555 
Task: Task B11, PRS: 94.03 
Task: Task B12, PRS: 94.022 
Task: Task C11, PRS: 7.0 
Task: Task C12, PRS: 6.999 
Task: Task C24, PRS: -399.847 
Task: Task C21, PRS: -399.887 
Task: Task C23, PRS: -400.043 
Task: Task C22, PRS: -402.878 
Net PR Sum: 3510
\end{verbatim}}
\item{
\begin{verbatim}
Tasks Completed:  ['Task A12', 'Task A13', 'Task A11', 'Task A23', 'Task A22', 'Task A21', 
'Task B21']
Task: Task B25, PRS: 964.499 
Task: Task B23, PRS: 964.499 
Task: Task B22, PRS: 964.496 
Task: Task B24, PRS: 963.76 
Task: Task B11, PRS: 94.03 
Task: Task B12, PRS: 94.022 
Task: Task C11, PRS: 7.0 
Task: Task C12, PRS: 6.999 
Task: Task C24, PRS: -399.847 
Task: Task C21, PRS: -399.887 
Task: Task C23, PRS: -400.043 
Task: Task C22, PRS: -402.878 
Net PR Sum: 2457
\end{verbatim}}
\item{
\begin{verbatim}
Tasks Completed:  ['Task A12', 'Task A13', 'Task A11', 'Task A23', 'Task A22', 'Task A21', 
'Task B21', 'Task B25']
Task: Task B23, PRS: 946.688 
Task: Task B22, PRS: 946.687 
Task: Task B24, PRS: 945.961 
Task: Task B11, PRS: 94.03 
Task: Task B12, PRS: 94.022 
Task: Task C11, PRS: 7.0 
Task: Task C12, PRS: 6.999 
Task: Task C24, PRS: -399.847 
Task: Task C21, PRS: -399.887 
Task: Task C23, PRS: -400.043 
Task: Task C22, PRS: -402.878 
Net PR Sum: 1439
\end{verbatim}}
\item{
\begin{verbatim}
Tasks Completed:  ['Task A12', 'Task A13', 'Task A11', 'Task A23', 'Task A22', 'Task A21', 
'Task B21', 'Task B25', 'Task B23']
Task: Task B22, PRS: 937.782 
Task: Task B24, PRS: 937.06 
Task: Task B11, PRS: 94.03 
Task: Task B12, PRS: 94.022 
Task: Task C11, PRS: 7.0 
Task: Task C12, PRS: 6.999 
Task: Task C24, PRS: -399.847 
Task: Task C21, PRS: -399.887 
Task: Task C23, PRS: -400.043 
Task: Task C22, PRS: -402.878 
Net PR Sum: 474
\end{verbatim}}
\item{
\begin{verbatim}
Tasks Completed:  ['Task A12', 'Task A13', 'Task A11', 'Task A23', 'Task A22', 'Task A21', 
'Task B21', 'Task B25', 'Task B23', 'Task B22']
Task: Task B24, PRS: 929.968 
Task: Task B11, PRS: 94.03 
Task: Task B12, PRS: 94.022 
Task: Task C11, PRS: 7.0 
Task: Task C12, PRS: 6.999 
Task: Task C24, PRS: -399.847 
Task: Task C21, PRS: -399.887 
Task: Task C23, PRS: -400.043 
Task: Task C22, PRS: -402.878 
Net PR Sum: -471
\end{verbatim}}
\item{
\begin{verbatim}
Tasks Completed:  ['Task A12', 'Task A13', 'Task A11', 'Task A23', 'Task A22', 'Task A21', 
'Task B21', 'Task B25', 'Task B23', 'Task B22', 'Task B24']
Task: Task B12, PRS: 543.896 
Task: Task B11, PRS: 543.814 
Task: Task C11, PRS: 7.0 
Task: Task C12, PRS: 6.999 
Task: Task C24, PRS: -399.847 
Task: Task C21, PRS: -399.887 
Task: Task C23, PRS: -400.043 
Task: Task C22, PRS: -402.878 
Net PR Sum: -501
    
\end{verbatim}}
\item{
\begin{verbatim}
Tasks Completed:  ['Task A12', 'Task A13', 'Task A11', 'Task A23', 'Task A22', 'Task A21', 
'Task B21', 'Task B25', 'Task B23', 'Task B22', 'Task B24', 'Task B12']
Task: Task B11, PRS: 526.001 
Task: Task C11, PRS: 7.0 
Task: Task C12, PRS: 6.999 
Task: Task C24, PRS: -399.847 
Task: Task C21, PRS: -399.887 
Task: Task C23, PRS: -400.043 
Task: Task C22, PRS: -402.878 
Net PR Sum: -1063
\end{verbatim}}
\item{
\begin{verbatim}
Tasks Completed:  ['Task A12', 'Task A13', 'Task A11', 'Task A23', 'Task A22', 'Task A21', 
'Task B21', 'Task B25', 'Task B23', 'Task B22', 'Task B24', 'Task B12', 'Task B11']
Task: Task C11, PRS: 7.0 
Task: Task C12, PRS: 6.999 
Task: Task C24, PRS: -399.847 
Task: Task C21, PRS: -399.887 
Task: Task C23, PRS: -400.043 
Task: Task C22, PRS: -402.878 
Net PR Sum: -1589
\end{verbatim}}
\end{itemize}

\subsection{Comparison with exact solution}\label{sec:compare}
As shown in Section~\ref{sec:specific-eg}, we assume that the user always selects a task with the highest points in the incentivised to-do list until the points for the the task with the highest points is less than the slack-off reward. This strategy is called the myopic greedy strategy. The performance of the myopic greedy strategy on to-do lists incentivized by the points computed by our proposed algorithm is compared with its performance on the same to-do lists when they are incentivized by the exact point values calculated using Value Iteration \cite{bellman1957markovian}. The exact method can be understood as applying Algorithm~\ref{alg:solvenext} and Algorithm~\ref{alg:expReward} in a flattened SMDP where the action space and state consists of actions which correspond to the completion of every task in the to-do list. The performance metric selected is the actual aggregated reward. The actual reward consists of the reward for completing a task (intrinsic reward) and the reward for completing a goal (value associated to goal). In case the deadline of a task or goal is missed, the value associated to the completion of the task or goal is not included.

We define a loss ratio (lr) metric to compare the performance of the proposed algorithm with the exact solution as follows:
\begin{equation}
    \mathrm{lr} = 100 \cdot \frac{ (R_{\mathrm{exact}} - R_{\mathrm{algorithm}})}{R_{\mathrm{exact}}},
\end{equation}
where $R_{\mathrm{exact}}$ and $R_{\mathrm{algorithm}}$ are the returns that our model of a myopic worker achieves when the to-do list is incentivized by the points computed with the exact method and our new approximate method, respectively.

\begin{figure}[h]
        \centering
        \includegraphics[width=0.95\textwidth]{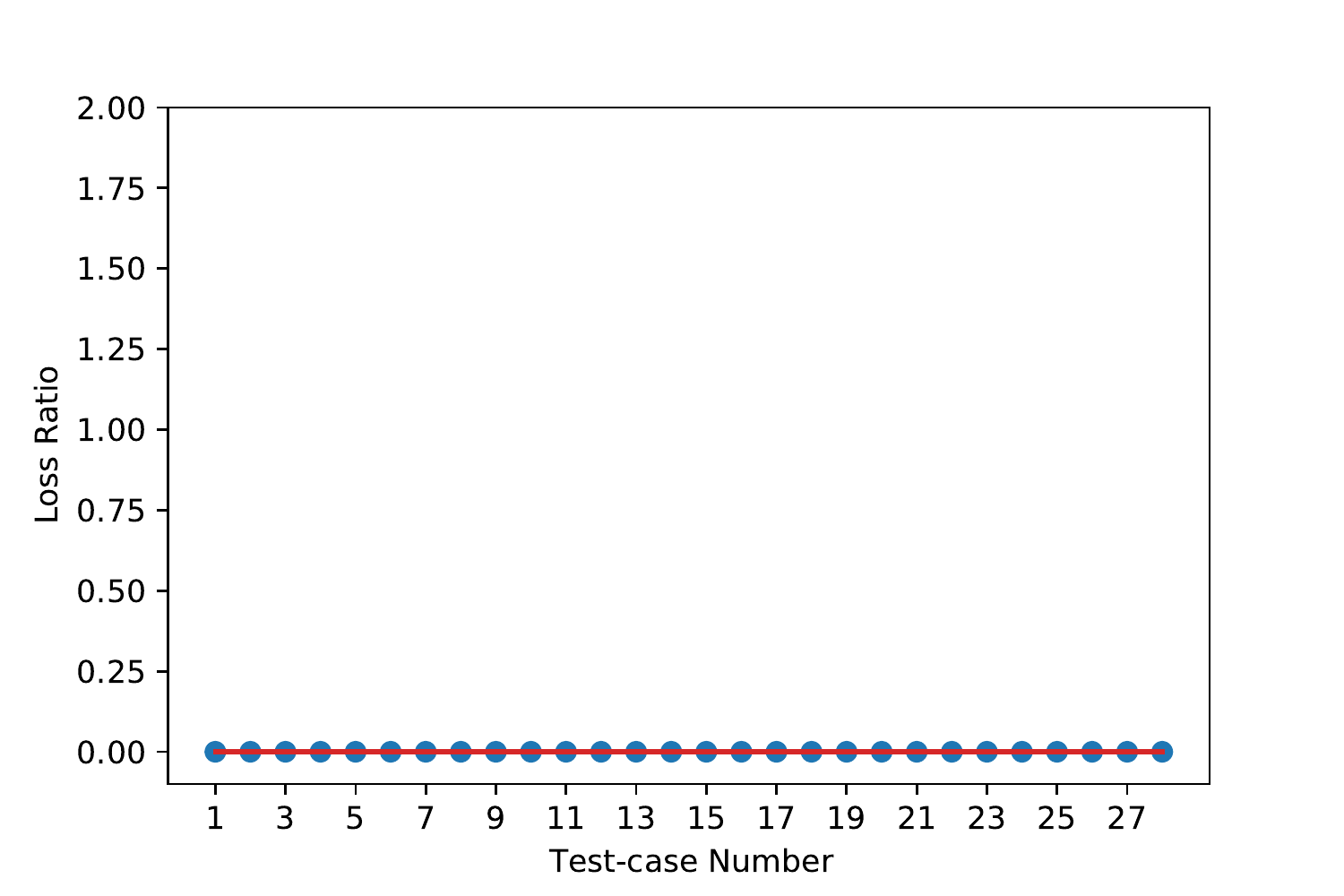}
        \caption{Loss ratio for the $28$ case studies. The number on the x-axis denotes the test-case as listed in the list of case studies described above in Appendix~\ref{sec:testcases}}
        \label{fig:lr-testcases}
    \end{figure}

In Figure~\ref{fig:lr-testcases}, the loss ratio has been plotted for $28$ hand-crafted case studies. These case studies are designed to cover most of the scenarios possible to study the algorithm's performance. The $28$ case studies are described in detail in Appendix~\ref{sec:testcases}.

Figure~\ref{fig:lr-testcases} depicts the loss-ratio of the $28$ case studies. A loss-ratio of $0$ indicates that the performance of the sequence of tasks followed by using the myopic greedy strategy of the optimal gamified points computed by our proposed algorithm and the exact solution is the same. All of the $28$ case studies yield a $0$ loss-ratio which indicates that the calculation of the gamified points using our proposal yields the exact same performance as using the optimal gamified points.
    
\subsection{Scalability of algorithm}\label{sec:scale}
While there is no theoretical limit for the size of the to-do list for which the proposed algorithm can solve, there is a practical limit set by the services used to run the API. The API is hosted on a Heroku server which has a practical $30$ seconds time limit for the API request to be active. We compared the time required for solving a to-do list in its initial state with varying number of goals, maximum depth and the branching factor of the to-do list. The maximum depth of the to-do list is defined as the lowest level a task can be abstractized. The branching factor is the number of sub-goals a goal can be divided into. For example, the maximum depth of the to-do list illustrated in Figure~\ref{fig:graph} is $3$ with a branching factor of $3$. The tasks generated in the scalability assessment are all essential, require an estimated time of $1$ time unit to be completed, have an intrinsic reward of $1$ and and have equal importance to other tasks.

\begin{figure}[h]
        \centering
        \includegraphics[width=0.9\textwidth]{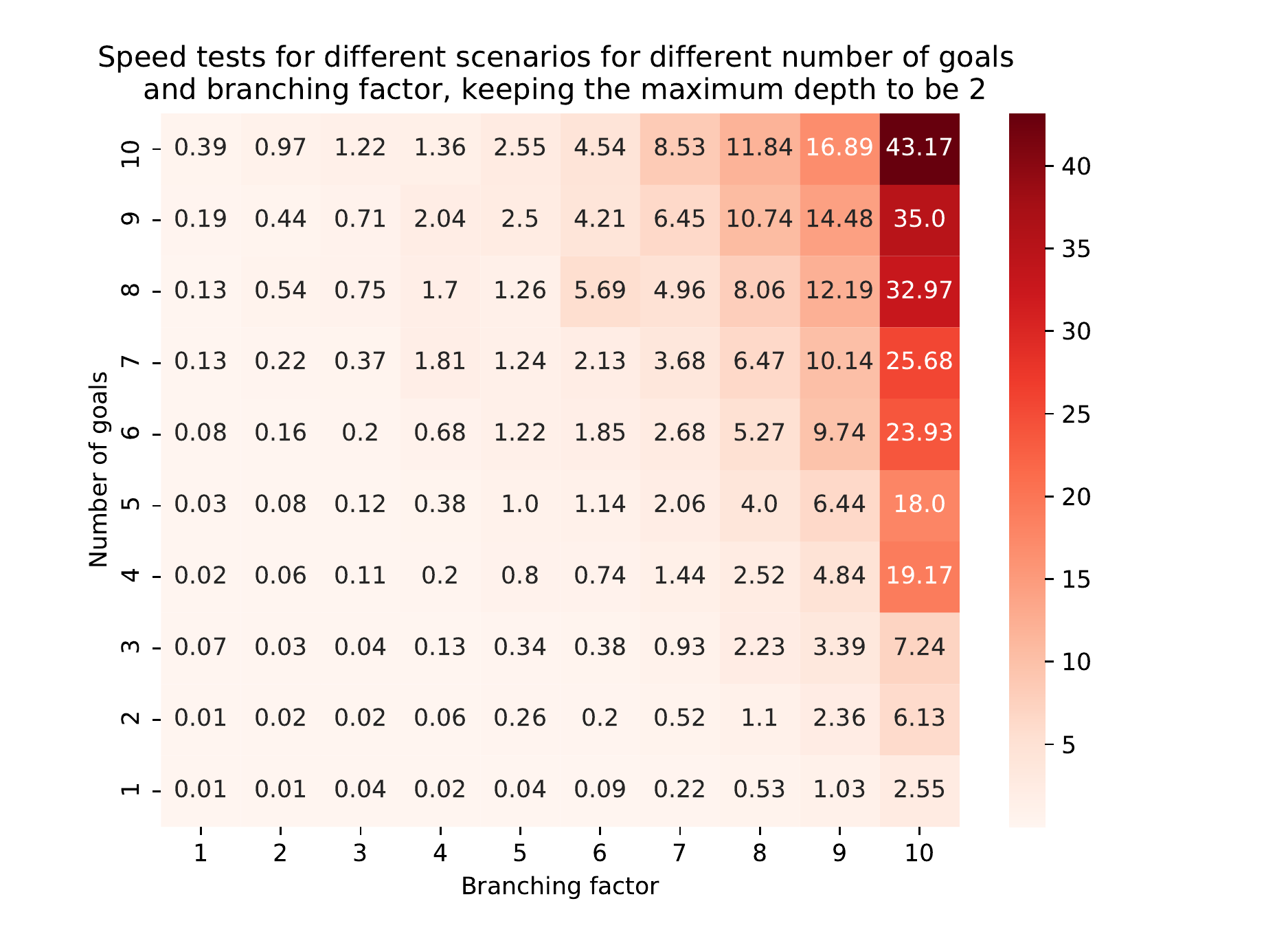}
        \caption{Heat-map representing the time taken (seconds) for case studies with varying number of goals and branching factors with a maximum depth of $2$}
        \label{fig:heatmap-goal-bf}
\end{figure}

\begin{figure}[H]
        \centering
        \includegraphics[width=0.9\textwidth]{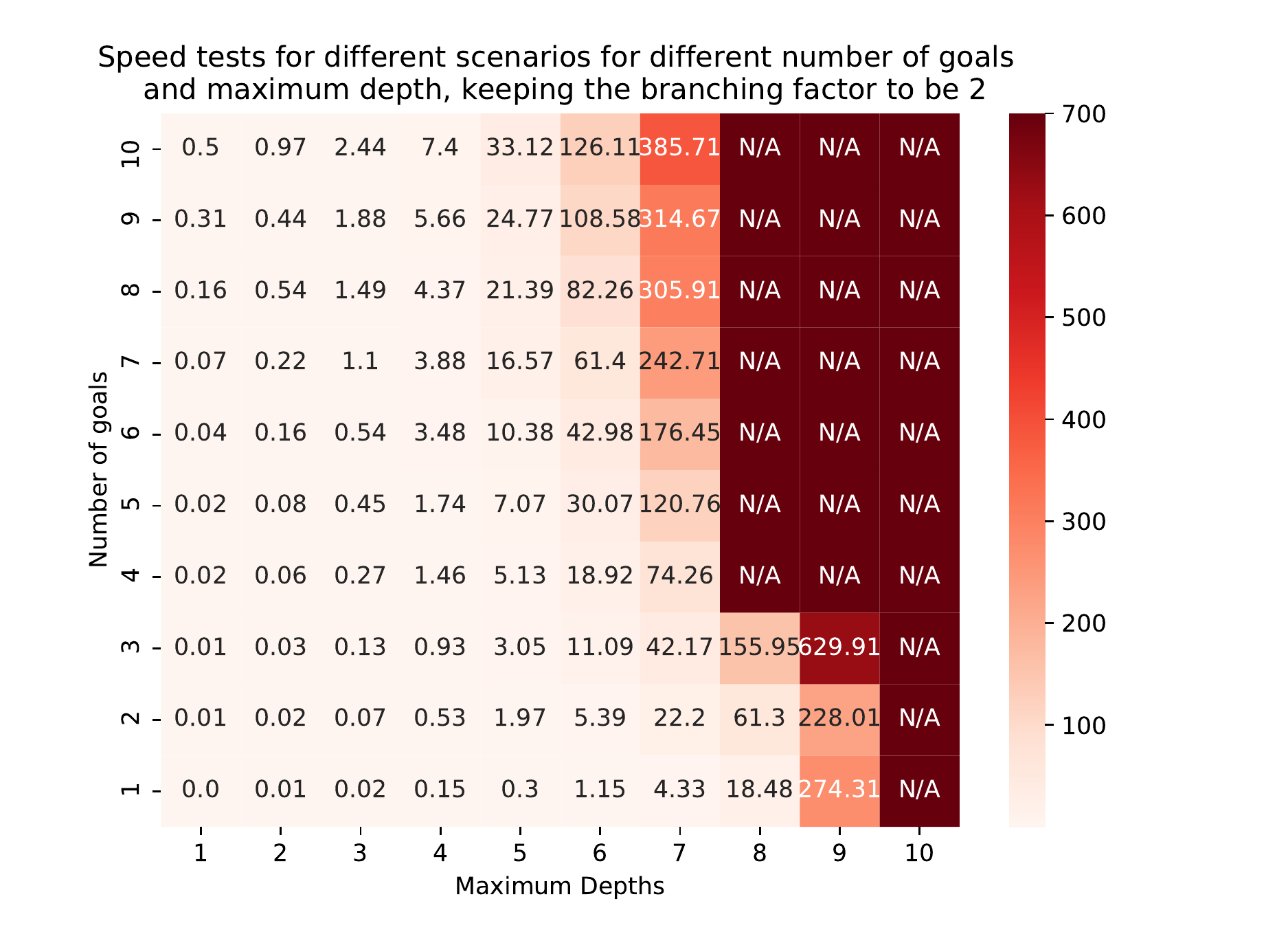}
        \caption{Heat-map representing the time taken (seconds) for case studies with varying number of goals and maximum depths with a branching factor of $2$}
        \label{fig:heatmap-goal-d}
\end{figure}

\begin{figure}[H]
        \centering
        \includegraphics[width=0.9\textwidth]{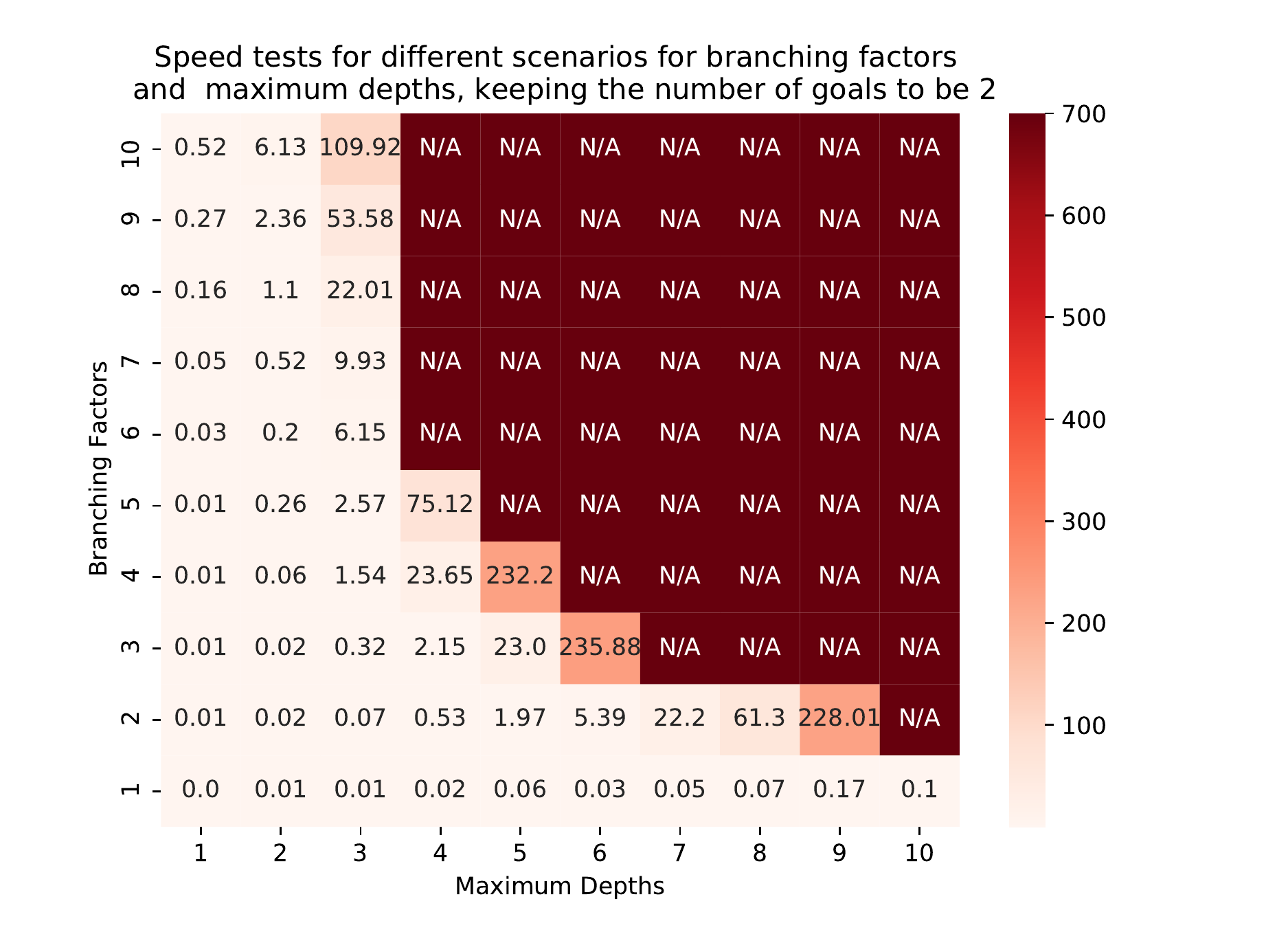}
        \caption{Heat-map representing the time taken (seconds) for case studies with varying number of branching factors and maximum depths and $2$ goals}
        \label{fig:heatmap-bf-d}
\end{figure}

While Figure~\ref{fig:heatmap-goal-bf} shows that the time required scales linearly to the increase in the number of goals, Figure~\ref{fig:heatmap-goal-d} and Figure~\ref{fig:heatmap-bf-d} show that the time required grows exponentially to the increase in branching factor and maximum depth. 
    
\begin{figure}[H]
        \centering
        \includegraphics[width=0.9\textwidth]{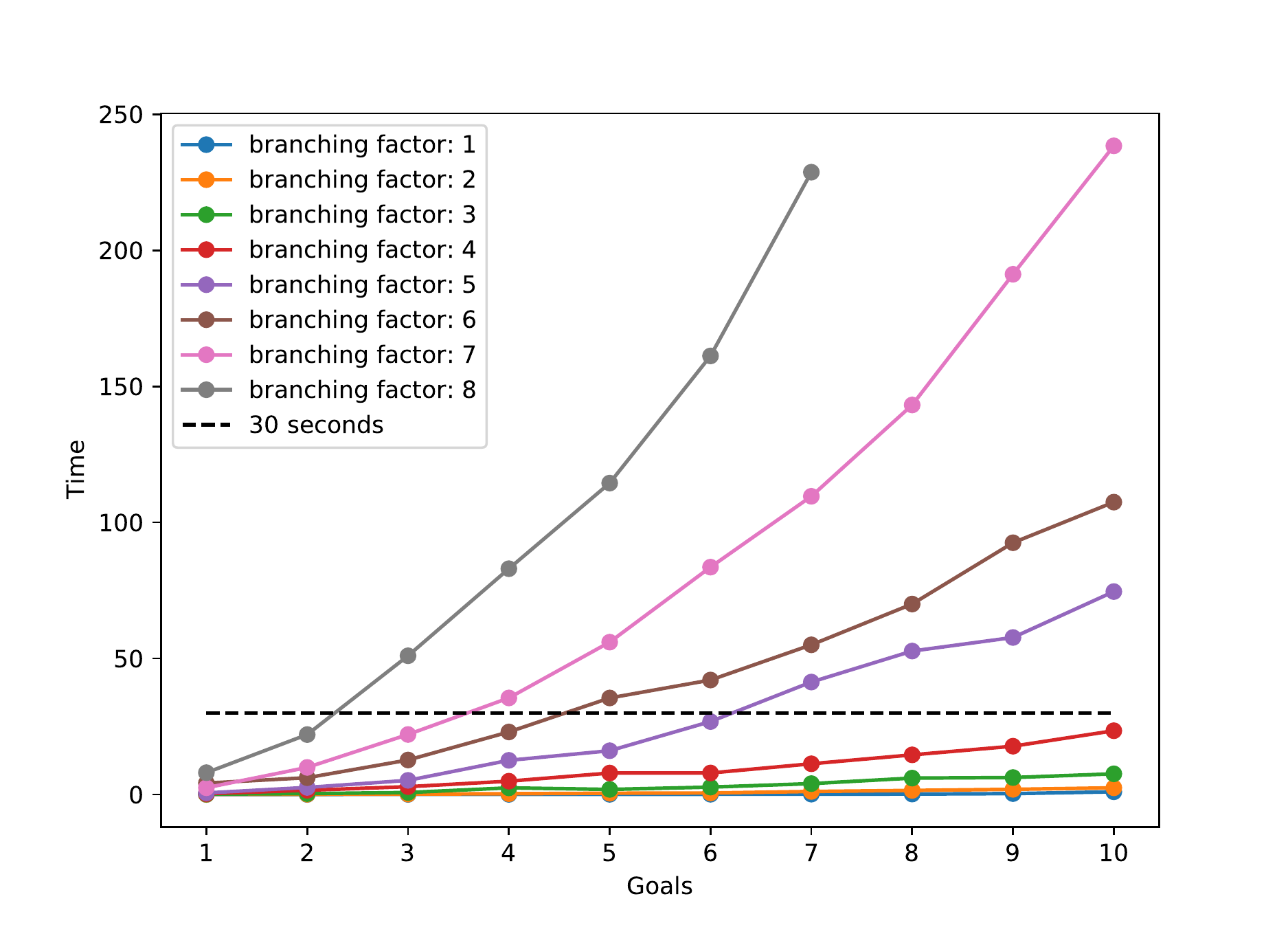}
        \caption{Speed-test to see the trend of increasing goals for a given branching factor with a maximum depth of $3$}
        \label{fig:goal_trend}
\end{figure}

Even with such constraints, Figure~\ref{fig:goal_trend} shows that a to-do list with a depth of 3 and branching factor of 4 is easily solvable by our proposed algorithm. Such a to-do list has a total of $\mathbf{576}$ tasks, which is more than big enough for most real-life examples. This shows that our proposed algorithm is practically useful.

\section{API documentation}\label{sec:API}
The API that we have developed can serve as a back-end to any to-do list gamification application.
It is available online at \url{https://github.com/RationalityEnhancement/todolistAPI/tree/multi_smdp_points}.
A demonstration of the API in action can be seen at \url{https://saksham36.github.io/todolistAPI-demoWebsite/}.

The communication between the gamification application and the API occurs in the following manner:
\begin{enumerate}
    \item  The gamification application sends a POST request with a specific URL to the API hosted
on Heroku, which provides information about the to-do list in JSON format. Details on the format expected as input are provided in Section~\ref{sec:api-input}.
\item After the POST request is received, the API parses the provided to-do-list information, computes task incentives, proposes a daily schedule of incentivized tasks, and sends this information back to the gamification application in JSON format. Details on the output that the API returns are provided in Section~\ref{sec:api-output}
\end{enumerate}

Additionally, the API requires a MongoDB database in order to be able to store the information generated by the API. For details on each of these items, please refer to the \href{https://github.com/RationalityEnhancement/todolistAPI/blob/multi_smdp_points/README.md}{\texttt{README.md}} file of the repository.

\subsection{API inputs}\label{sec:api-input}
The API input consists of a POST request which consists of a Header and Body. The details of each are described in Section~\ref{sec:api-header} and Section~\ref{sec:api-body} respectively.
\subsubsection{Header}
The header consists of the URL which is sent to the API. The general pattern of the URL looks like: \\
\small{\inlinecode{http://<server>/api/<compulsoryParameters>/<additionalParameters>/tree/<userID>/<functionName>}}\normalsize.

\begin{itemize}
   \item \inlinecode{<server>}: DNS or IP address of the server. 
            \item \inlinecode{<compulsoryParameters>}: Parameters that are \textit{independent} of the incentivizing method.
            \item \inlinecode{<additionalParameters>}: Parameters that are related to the incentivizing method.
            \item \inlinecode{<userID>}: A unique user identification code.
            \item \inlinecode{<functionName>}: The functionality that the API should provide.
\end{itemize}\label{sec:api-header}
\subsubsection{Body}
 Additionally, the request must contain a body in JSON format with the following information:
        \begin{itemize}
            \item \inlinecode{currentIntentionsList}: List of tasks that have already been scheduled. Each item in this list represents a scheduled task and it has to contain the following information:
            \begin{itemize}
                \item \inlinecode{\_c}: Goal code/number.
                \item \inlinecode{\_id}: Unique identification code of the scheduled task.
                \item \inlinecode{d}: Whether the scheduled task has been completed or not.
                \item \inlinecode{nvm}: Whether the scheduled task has been marked to be completed at some other time.
                \item \inlinecode{t}: Title of the scheduled task.
                \item \inlinecode{vd}: Value of the scheduled task.
            \end{itemize}
            \item \inlinecode{projects}: Tree of goals and their respective tasks. Each item (goal or task) is composed of the following information:
            \begin{itemize}
                \item \inlinecode{id}: Unique identification code of the item.
                \item \inlinecode{nm}: Title of the item.
                \item \inlinecode{lm}: Time stamp of item's last modification.
                \item \inlinecode{cp}: Time stamp of item's completion.
                \item \inlinecode{ch}: Sub-items of the current item.
            \end{itemize}
            
            \item \inlinecode{timezoneOffsetMinutes}: Time difference in minutes between user's time zone and UTC.
            \item \inlinecode{today\_hours}: Number of hours that a user would like to work on the current day (today).
            \item \inlinecode{typical\_hours}: Number of hours that a user would like to work on a typical day. m
            \item \inlinecode{userkey}: Unique identification code of the user.
            \item \inlinecode{updated}: Time stamp of the last modification of the items in the \inlinecode{projects} tree.
        \end{itemize}
        Each to-do-list item (i.e. goal or task) title follows patterns that encode all the necessary information. The following list describes these patterns in detail:
        \begin{itemize}
            \item \inlinecode{\#CG<N>\_<goal\_name>} defines a goal name, where \inlinecode{<N>} is the number of the goal and \inlinecode{<goal\_name>} is the actual goal name specified by the user.
            \item \inlinecode{==<value>} defines a value of a goal/tasks, where \inlinecode{<value>} $\in \mathbb{Z}_{\ge 0}$.
            \item \inlinecode{$\sim\sim$<time\_estimate> <time\_unit>} defines a time estimate for a task, where \inlinecode{<time\_estimate> min} $\in \mathbb{N}$ corresponds to the number of minutes \textit{or} \inlinecode{<time\_estimate> h} $\in \mathbb{R}_{> 0}$ corresponds to the amount of hours.
            \item \inlinecode{DUE:<YYYY-MM-DD> <HH:mm>} defines a deadline, where \inlinecode{<YYYY-MM-DD>} defines a date according to the ISO format and \inlinecode{<HH:mm>} defines a 24-hours day time. If \inlinecode{<HH:mm>} is not provided, then 23:59 is taken as a default day-time value.
            \item \inlinecode{IMPORTANCE: <importance>} defines the importance for completing the task/sub-goal for the completion its super-ordinate goal, where \inlinecode{<importance>} $\in \mathbb{Z}_{\ge 0}$.
            \item \inlinecode{Intrinsic Value: <intrinsic value>} defines the value for completion of the task/sub-goal is independent of its super-ordinate goal, where \inlinecode{<intrinsic value} $\in \mathbb{Z}_{\ge 0}$.
            \item \inlinecode{Essential:: <essential>} is a boolean value which is true if the task/sub-goal is essential for the completion of the task independent of its super-ordinate goal, 
            \item \inlinecode{\#HOURS\_TYPICAL ==<hours>} defines the total number of hours per day, i.e. the amount of hours $\in (0, 24]$ that a user wants to work on a typical day.
            \item \inlinecode{\#HOURS\_TODAY ==<hours>} defines the total number of hours for today, i.e. the amount of hours $\in (0, 24]$ that a user wants to work today.
            \item Scheduling tags that users can accompany to their tasks:
            \begin{itemize}
                \item \inlinecode{\#daily} represents a task that is repetitive on a daily basis.
                \item \inlinecode{\#future} represents a task that has to be scheduled at some point in the future, but not at the moment.
                \item \inlinecode{\#today} represents a task that has to be scheduled today.
                \item \inlinecode{\#<weekday>} represents a task that has to be scheduled on a specific weekday (where \inlinecode{weekday} is a day from Monday to Sunday). If this task is repetitive on a weekly basis, a plural suffix is appended to the same tag, i.e. \inlinecode{\#<weekday>s}.
                \item \inlinecode{\#weekdays} represents a repetitive task that has to be scheduled on each working day (from Monday to Friday).
                \item \inlinecode{\#weekends} represents a repetitive task that has to be scheduled during weekends (Saturday and Sunday).
                \item \inlinecode{\#YYYY-MM-DD} represents a task that has to be scheduled on a specific day according to the ISO standard (year-month-day).
            \end{itemize}
        \end{itemize}\label{sec:api-body}
        
\subsection{API outputs}\label{sec:api-output}
 After generating incentives for each task in a to-do-list, the API selects a subset of them and it proposes an incentivized daily schedule as output. The output is a list of dictionaries in JSON format and it contains the following information for each task in the list:
        \begin{itemize}
            \item \inlinecode{id}: Unique identification code of the task.
            \item \inlinecode{nm}: Human-readable name of the task.
            \item \inlinecode{lm}: Time stamp of task's last modification.
            \item \inlinecode{est}: Time estimate of the task.
            \item \inlinecode{parentId}: Unique identification code of the goal which the task belongs to.
            \item \inlinecode{pcp}: Whether the parent node (i.e. goal) has been completed.
            \item \inlinecode{val}: Generated incentive for the task.
        \end{itemize}

\section{Future Work}\label{sec:future}
We consider multiple potential ways to improve the API in order to make its functionality even
closer to real-world demands. Usability enhancements will include supporting tasks
that contribute to multiple goals simultaneously. Additionally, decreasing the time complexity for generating solutions for larger to-do lists is of the highest priority. One possible future improvement would be to produce fast responses after making minor
changes in the input information even if the changes may modify the solution. Additionally, work on getting better time estimates fot the planning-fallacy constant would enable for more precise incentivization.

\section*{Acknowledgment}
This work was supported by grant number 1757269 from the National Science Foundation.
\newpage
\bibliographystyle{plain}
\bibliography{ref.bib}
\newpage
\input{appendix}

\end{document}

%% file: appendix.tex
\addcontentsline{toc}{section}{Appendix}
\section*{Appendix}  \label{sec:appendix}
\renewcommand{\thesubsection}{A\arabic{subsection}}
\subsection{Case studies for comparison with exact solution}\label{sec:testcases}
The following lists the $28$ case studies used in Section~\ref{sec:compare}. For every case study, each goal is assigned a \texttt{Value} which represents the value associated for completion of a goal, similarly, there is a \texttt{Deadline} which indicates how many time-units the user has to complete the goal. Each sub-goal has an \texttt{Intrinsic Reward} which denotes the reward derived of completing the sub-goal, independent of the corresponding goal. \texttt{Ess} is a Boolean value which is \texttt{True} is the sub-goal is deemed essential to complete the corresponding goal. Additionally, 
\texttt{Imp} denotes how important the sub-goal is relative to the other sub-goals. Finally, \texttt{Time} indicates the estimated time to complete a sub-goal.
\begin{enumerate}
    \item{1 goal with 2 sub-goals, with varying importance}
\begin{verbatim}
 Goal A - Value: 100, Deadline: 12

     SG A1 - Intrinsic Reward: 0, Ess: True,Imp: 90, Time: 3

        Task A11 - Intrinsic Reward: 0, Ess: True,Imp: 30, Time: 1

        Task A12 - Intrinsic Reward: 0, Ess: True,Imp: 30, Time: 1

        Task A13 - Intrinsic Reward: 0, Ess: True,Imp: 30, Time: 1

     SG A2 - Intrinsic Reward: 0, Ess: True,Imp: 10, Time: 3

        Task A21 - Intrinsic Reward: 0, Ess: True,Imp: 3, Time: 1

        Task A22 - Intrinsic Reward: 0, Ess: True,Imp: 3, Time: 1

        Task A23 - Intrinsic Reward: 0, Ess: True,Imp: 4, Time: 1
\end{verbatim}
\item{3 goals with 2 sub-goals each, in a realistic example}
\begin{verbatim}
 Goal A - Value: 1000, Deadline: 12

     SG A1 - Intrinsic Reward: 40, Ess: True,Imp: 100, Time: 7

        Task A11 - Intrinsic Reward: 10, Ess: True,Imp: 60, Time: 3

        Task A12 - Intrinsic Reward: 15, Ess: True,Imp: 20, Time: 2

        Task A13 - Intrinsic Reward: 15, Ess: True,Imp: 20, Time: 2

     SG A2 - Intrinsic Reward: 30, Ess: True,Imp: 100, Time: 5

        Task A21 - Intrinsic Reward: 20, Ess: True,Imp: 60, Time: 3

        Task A22 - Intrinsic Reward: 2, Ess: True,Imp: 30, Time: 1

        Task A23 - Intrinsic Reward: 8, Ess: True,Imp: 10, Time: 1

 Goal B - Value: 500, Deadline: 50

     SG B1 - Intrinsic Reward: 100, Ess: True,Imp: 100, Time: 6

        Task B11 - Intrinsic Reward: 80, Ess: True,Imp: 90, Time: 4

        Task B12 - Intrinsic Reward: 20, Ess: True,Imp: 10, Time: 2

     SG B2 - Intrinsic Reward: 100, Ess: True,Imp: 100, Time: 17

        Task B21 - Intrinsic Reward: 20, Ess: True,Imp: 20, Time: 2

        Task B22 - Intrinsic Reward: 10, Ess: True,Imp: 60, Time: 2

        Task B23 - Intrinsic Reward: 10, Ess: True,Imp: 2, Time: 1

        Task B24 - Intrinsic Reward: 40, Ess: True,Imp: 15, Time: 10

        Task B25 - Intrinsic Reward: 20, Ess: True,Imp: 3, Time: 2

 Goal C - Value: 5000, Deadline: 50

     SG C1 - Intrinsic Reward: 10, Ess: True,Imp: 100, Time: 3

        Task C11 - Intrinsic Reward: 5, Ess: True,Imp: 60, Time: 1

        Task C12 - Intrinsic Reward: 5, Ess: True,Imp: 40, Time: 2

     SG C2 - Intrinsic Reward: 90, Ess: True,Imp: 100, Time: 502

        Task C21 - Intrinsic Reward: 50, Ess: True,Imp: 20, Time: 50

        Task C22 - Intrinsic Reward: 10, Ess: True,Imp: 60, Time: 400

        Task C23 - Intrinsic Reward: 20, Ess: True,Imp: 10, Time: 50

        Task C24 - Intrinsic Reward: 10, Ess: True,Imp: 10, Time: 2
\end{verbatim}
\item{Simple test-case with a single sub-goal}
\begin{verbatim}
 Goal A - Value: 100, Deadline: 12

     SG A1 - Intrinsic Reward: 0, Ess: True,Imp: 90, Time: 3

        Task A11 - Intrinsic Reward: 0, Ess: True,Imp: 30, Time: 1

        Task A12 - Intrinsic Reward: 0, Ess: True,Imp: 30, Time: 1

        Task A13 - Intrinsic Reward: 0, Ess: True,Imp: 30, Time: 1
\end{verbatim}
\item{1 goal with 2 sub-goals with uneven depth}
\begin{verbatim}
 Goal A - Value: 100, Deadline: 12

     SG A1 - Intrinsic Reward: 0, Ess: True,Imp: 90, Time: 3

        Task A11 - Intrinsic Reward: 0, Ess: True,Imp: 30, Time: 1

        Task A12 - Intrinsic Reward: 0, Ess: True,Imp: 30, Time: 1

        Task A13 - Intrinsic Reward: 0, Ess: True,Imp: 30, Time: 1

        Task A2 - Intrinsic Reward: 0, Ess: True,Imp: 10, Time: 1
\end{verbatim}
\item{1 goal with 2 sub-goals with varying intrinsic rewards}
\begin{verbatim}
 Goal A - Value: 100, Deadline: 12

     SG A1 - Intrinsic Reward: 5.0, Ess: True,Imp: 50, Time: 2

        Task A11 - Intrinsic Reward: 2.5, Ess: True,Imp: 25, Time: 1

        Task A12 - Intrinsic Reward: 2.5, Ess: True,Imp: 25, Time: 1

     SG A2 - Intrinsic Reward: 10, Ess: True,Imp: 50, Time: 2

        Task A21 - Intrinsic Reward: 5, Ess: True,Imp: 25, Time: 1

        Task A22 - Intrinsic Reward: 5, Ess: True,Imp: 25, Time: 1
\end{verbatim}
\item{1 goal with 2 sub-goals with varying essentialness}
\begin{verbatim}
 Goal A - Value: 100, Deadline: 12

     SG A1 - Intrinsic Reward: 0, Ess: True,Imp: 50, Time: 2

        Task A11 - Intrinsic Reward: 0, Ess: True,Imp: 25, Time: 1

        Task A12 - Intrinsic Reward: 0, Ess: True,Imp: 25, Time: 1

     SG A2 - Intrinsic Reward: 0, Ess: False,Imp: 50, Time: 2

        Task A21 - Intrinsic Reward: 0, Ess: False,Imp: 25, Time: 1

        Task A22 - Intrinsic Reward: 0, Ess: False,Imp: 25, Time: 1
\end{verbatim}
\item{1 goal with 2 sub-goals with varying essentialness with high probability of missing the deadline}
\begin{verbatim}
 Goal A - Value: 100, Deadline: 3

     SG A1 - Intrinsic Reward: 0, Ess: True,Imp: 50, Time: 2

        Task A11 - Intrinsic Reward: 0, Ess: True,Imp: 25, Time: 1

        Task A12 - Intrinsic Reward: 0, Ess: True,Imp: 25, Time: 1

     SG A2 - Intrinsic Reward: 0, Ess: False,Imp: 50, Time: 2

        Task A21 - Intrinsic Reward: 0, Ess: False,Imp: 25, Time: 1

        Task A22 - Intrinsic Reward: 0, Ess: False,Imp: 25, Time: 1
\end{verbatim}
\item{1 goal with 2 sub-goals with varying essentialness and intrinsic reward}
\begin{verbatim}
 Goal A - Value: 100, Deadline: 12

     SG A1 - Intrinsic Reward: 5.0, Ess: True,Imp: 50, Time: 2

        Task A11 - Intrinsic Reward: 2.5, Ess: True,Imp: 25, Time: 1

        Task A12 - Intrinsic Reward: 2.5, Ess: True,Imp: 25, Time: 1

     SG A2 - Intrinsic Reward: 10, Ess: False,Imp: 50, Time: 2

        Task A21 - Intrinsic Reward: 5, Ess: False,Imp: 25, Time: 1

        Task A22 - Intrinsic Reward: 5, Ess: False,Imp: 25, Time: 1
\end{verbatim}
\item{1 goal with 2 sub-goals with varying essentialness and intrinsic reward with high probability of missing the deadline}
\begin{verbatim}
 Goal A - Value: 100, Deadline: 3

     SG A1 - Intrinsic Reward: 5.0, Ess: True,Imp: 50, Time: 2

        Task A11 - Intrinsic Reward: 2.5, Ess: True,Imp: 25, Time: 1

        Task A12 - Intrinsic Reward: 2.5, Ess: True,Imp: 25, Time: 1

     SG A2 - Intrinsic Reward: 10, Ess: False,Imp: 50, Time: 2

        Task A21 - Intrinsic Reward: 5, Ess: False,Imp: 25, Time: 1

        Task A22 - Intrinsic Reward: 5, Ess: False,Imp: 25, Time: 1
\end{verbatim}
\item{1 goal with 2 sub-goals with varying essentialness and importance split of 60-40}
\begin{verbatim}
 Goal A - Value: 100, Deadline: 12

     SG A1 - Intrinsic Reward: 0, Ess: True,Imp: 60, Time: 2

        Task A11 - Intrinsic Reward: 0, Ess: True,Imp: 30, Time: 1

        Task A12 - Intrinsic Reward: 0, Ess: True,Imp: 30, Time: 1

     SG A2 - Intrinsic Reward: 0, Ess: False,Imp: 40, Time: 2

        Task A21 - Intrinsic Reward: 0, Ess: False,Imp: 20, Time: 1

        Task A22 - Intrinsic Reward: 0, Ess: False,Imp: 20, Time: 1
\end{verbatim}
\item{1 goal with 2 sub-goals with varying essentialness and importance split of 60-40 with high probability of missing the deadline}
\begin{verbatim}
 Goal A - Value: 100, Deadline: 3

     SG A1 - Intrinsic Reward: 0, Ess: True,Imp: 60, Time: 2

        Task A11 - Intrinsic Reward: 0, Ess: True,Imp: 30, Time: 1

        Task A12 - Intrinsic Reward: 0, Ess: True,Imp: 30, Time: 1

     SG A2 - Intrinsic Reward: 0, Ess: False,Imp: 40, Time: 2

        Task A21 - Intrinsic Reward: 0, Ess: False,Imp: 20, Time: 1

        Task A22 - Intrinsic Reward: 0, Ess: False,Imp: 20, Time: 1
\end{verbatim}
\item{1 goal with 2 sub-goals with varying essentialness, intrinsic reward, and importance split of 60-40}
\begin{verbatim}
 Goal A - Value: 100, Deadline: 12

     SG A1 - Intrinsic Reward: 5.0, Ess: True,Imp: 60, Time: 2

        Task A11 - Intrinsic Reward: 2.5, Ess: True,Imp: 30, Time: 1

        Task A12 - Intrinsic Reward: 2.5, Ess: True,Imp: 30, Time: 1

     SG A2 - Intrinsic Reward: 10, Ess: False,Imp: 40, Time: 2

        Task A21 - Intrinsic Reward: 5, Ess: False,Imp: 20, Time: 1

        Task A22 - Intrinsic Reward: 5, Ess: False,Imp: 20, Time: 1
\end{verbatim}
\item{1 goal with 2 sub-goals with varying essentialness, intrinsic reward, and importance split of 60-40 with high probability of missing the deadline}
\begin{verbatim}
 Goal A - Value: 100, Deadline: 3

     SG A1 - Intrinsic Reward: 5.0, Ess: True,Imp: 60, Time: 2

        Task A11 - Intrinsic Reward: 2.5, Ess: True,Imp: 30, Time: 1

        Task A12 - Intrinsic Reward: 2.5, Ess: True,Imp: 30, Time: 1

     SG A2 - Intrinsic Reward: 10, Ess: False,Imp: 40, Time: 2

        Task A21 - Intrinsic Reward: 5, Ess: False,Imp: 20, Time: 1

        Task A22 - Intrinsic Reward: 5, Ess: False,Imp: 20, Time: 1
\end{verbatim}
\item{1 goal with 2 sub-goals with varying essentialness and importance split of 70-30}
\begin{verbatim}
 Goal A - Value: 100, Deadline: 12

     SG A1 - Intrinsic Reward: 0, Ess: True,Imp: 70, Time: 2

        Task A11 - Intrinsic Reward: 0, Ess: True,Imp: 35, Time: 1

        Task A12 - Intrinsic Reward: 0, Ess: True,Imp: 35, Time: 1

     SG A2 - Intrinsic Reward: 0, Ess: False,Imp: 30, Time: 2

        Task A21 - Intrinsic Reward: 0, Ess: False,Imp: 15, Time: 1

        Task A22 - Intrinsic Reward: 0, Ess: False,Imp: 15, Time: 1
\end{verbatim}
\item{1 goal with 2 sub-goals with varying essentialness and importance split of 70-30 with high probability of missing the deadline}
\begin{verbatim}
 Goal A - Value: 100, Deadline: 3

     SG A1 - Intrinsic Reward: 0, Ess: True,Imp: 70, Time: 2

        Task A11 - Intrinsic Reward: 0, Ess: True,Imp: 35, Time: 1

        Task A12 - Intrinsic Reward: 0, Ess: True,Imp: 35, Time: 1

     SG A2 - Intrinsic Reward: 0, Ess: False,Imp: 30, Time: 2

        Task A21 - Intrinsic Reward: 0, Ess: False,Imp: 15, Time: 1

        Task A22 - Intrinsic Reward: 0, Ess: False,Imp: 15, Time: 1
\end{verbatim}
\item{1 goal with 2 sub-goals with varying essentialness, intrinsic reward, and importance split of 70-30}
\begin{verbatim}
 Goal A - Value: 100, Deadline: 12

     SG A1 - Intrinsic Reward: 5.0, Ess: True,Imp: 70, Time: 2

        Task A11 - Intrinsic Reward: 2.5, Ess: True,Imp: 35, Time: 1

        Task A12 - Intrinsic Reward: 2.5, Ess: True,Imp: 35, Time: 1

     SG A2 - Intrinsic Reward: 10, Ess: False,Imp: 30, Time: 2

        Task A21 - Intrinsic Reward: 5, Ess: False,Imp: 15, Time: 1

        Task A22 - Intrinsic Reward: 5, Ess: False,Imp: 15, Time: 1
\end{verbatim}
\item{1 goal with 2 sub-goals with varying essentialness, intrinsic reward, and importance split of 70-30 with high probability of missing the deadline}
\begin{verbatim}
 Goal A - Value: 100, Deadline: 3

     SG A1 - Intrinsic Reward: 5.0, Ess: True,Imp: 70, Time: 2

        Task A11 - Intrinsic Reward: 2.5, Ess: True,Imp: 35, Time: 1

        Task A12 - Intrinsic Reward: 2.5, Ess: True,Imp: 35, Time: 1

     SG A2 - Intrinsic Reward: 10, Ess: False,Imp: 30, Time: 2

        Task A21 - Intrinsic Reward: 5, Ess: False,Imp: 15, Time: 1

        Task A22 - Intrinsic Reward: 5, Ess: False,Imp: 15, Time: 1
\end{verbatim}
\item{1 goal with 2 sub-goals with varying essentialness and importance split of 80-20}
\begin{verbatim}
 Goal A - Value: 100, Deadline: 12

     SG A1 - Intrinsic Reward: 0, Ess: True,Imp: 80, Time: 2

        Task A11 - Intrinsic Reward: 0, Ess: True,Imp: 40, Time: 1

        Task A12 - Intrinsic Reward: 0, Ess: True,Imp: 40, Time: 1

     SG A2 - Intrinsic Reward: 0, Ess: False,Imp: 20, Time: 2

        Task A21 - Intrinsic Reward: 0, Ess: False,Imp: 10, Time: 1

        Task A22 - Intrinsic Reward: 0, Ess: False,Imp: 10, Time: 1
\end{verbatim}
\item{1 goal with 2 sub-goals with varying essentialness and importance split of 80-20 with high probability of missing the deadline}
\begin{verbatim}
 Goal A - Value: 100, Deadline: 3

     SG A1 - Intrinsic Reward: 0, Ess: True,Imp: 80, Time: 2

        Task A11 - Intrinsic Reward: 0, Ess: True,Imp: 40, Time: 1

        Task A12 - Intrinsic Reward: 0, Ess: True,Imp: 40, Time: 1

     SG A2 - Intrinsic Reward: 0, Ess: False,Imp: 20, Time: 2

        Task A21 - Intrinsic Reward: 0, Ess: False,Imp: 10, Time: 1

        Task A22 - Intrinsic Reward: 0, Ess: False,Imp: 10, Time: 1
\end{verbatim}
\item{1 goal with 2 sub-goals with varying essentialness, intrinsic reward, and importance split of 80-20}
\begin{verbatim}
 Goal A - Value: 100, Deadline: 12

     SG A1 - Intrinsic Reward: 5.0, Ess: True,Imp: 80, Time: 2

        Task A11 - Intrinsic Reward: 2.5, Ess: True,Imp: 40, Time: 1

        Task A12 - Intrinsic Reward: 2.5, Ess: True,Imp: 40, Time: 1

     SG A2 - Intrinsic Reward: 10, Ess: False,Imp: 20, Time: 2

        Task A21 - Intrinsic Reward: 5, Ess: False,Imp: 10, Time: 1

        Task A22 - Intrinsic Reward: 5, Ess: False,Imp: 10, Time: 1
\end{verbatim}
\item{1 goal with 2 sub-goals with varying essentialness, intrinsic reward, and importance split of 80-20 with high probability of missing the deadline}
\begin{verbatim}
 Goal A - Value: 100, Deadline: 3

     SG A1 - Intrinsic Reward: 5.0, Ess: True,Imp: 80, Time: 2

        Task A11 - Intrinsic Reward: 2.5, Ess: True,Imp: 40, Time: 1

        Task A12 - Intrinsic Reward: 2.5, Ess: True,Imp: 40, Time: 1

     SG A2 - Intrinsic Reward: 10, Ess: False,Imp: 20, Time: 2

        Task A21 - Intrinsic Reward: 5, Ess: False,Imp: 10, Time: 1

        Task A22 - Intrinsic Reward: 5, Ess: False,Imp: 10, Time: 1
\end{verbatim}
\item{1 goal with 2 sub-goals with varying essentialness and importance split of 90-10}
\begin{verbatim}
 Goal A - Value: 100, Deadline: 12

     SG A1 - Intrinsic Reward: 0, Ess: True,Imp: 90, Time: 2

        Task A11 - Intrinsic Reward: 0, Ess: True,Imp: 45, Time: 1

        Task A12 - Intrinsic Reward: 0, Ess: True,Imp: 45, Time: 1

     SG A2 - Intrinsic Reward: 0, Ess: False,Imp: 10, Time: 2

        Task A21 - Intrinsic Reward: 0, Ess: False,Imp: 5, Time: 1

        Task A22 - Intrinsic Reward: 0, Ess: False,Imp: 5, Time: 1
\end{verbatim}
\item{1 goal with 2 sub-goals with varying essentialness and importance split of 90-10 with high probability of missing the deadline}
\begin{verbatim}
 Goal A - Value: 100, Deadline: 3

     SG A1 - Intrinsic Reward: 0, Ess: True,Imp: 90, Time: 2

        Task A11 - Intrinsic Reward: 0, Ess: True,Imp: 45, Time: 1

        Task A12 - Intrinsic Reward: 0, Ess: True,Imp: 45, Time: 1

     SG A2 - Intrinsic Reward: 0, Ess: False,Imp: 10, Time: 2

        Task A21 - Intrinsic Reward: 0, Ess: False,Imp: 5, Time: 1

        Task A22 - Intrinsic Reward: 0, Ess: False,Imp: 5, Time: 1
\end{verbatim}
\item{1 goal with 2 sub-goals with varying essentialness, intrinsic reward, and importance split of 90-10}
\begin{verbatim}
 Goal A - Value: 100, Deadline: 12

     SG A1 - Intrinsic Reward: 5.0, Ess: True,Imp: 90, Time: 2

        Task A11 - Intrinsic Reward: 2.5, Ess: True,Imp: 45, Time: 1

        Task A12 - Intrinsic Reward: 2.5, Ess: True,Imp: 45, Time: 1

     SG A2 - Intrinsic Reward: 10, Ess: False,Imp: 10, Time: 2

        Task A21 - Intrinsic Reward: 5, Ess: False,Imp: 5, Time: 1

        Task A22 - Intrinsic Reward: 5, Ess: False,Imp: 5, Time: 1
\end{verbatim}
\item{1 goal with 2 sub-goals with varying essentialness, intrinsic reward, and importance split of 90-10 with high probability of missing the deadline}
\begin{verbatim}
 Goal A - Value: 100, Deadline: 3

     SG A1 - Intrinsic Reward: 5.0, Ess: True,Imp: 90, Time: 2

        Task A11 - Intrinsic Reward: 2.5, Ess: True,Imp: 45, Time: 1

        Task A12 - Intrinsic Reward: 2.5, Ess: True,Imp: 45, Time: 1

     SG A2 - Intrinsic Reward: 10, Ess: False,Imp: 10, Time: 2

        Task A21 - Intrinsic Reward: 5, Ess: False,Imp: 5, Time: 1

        Task A22 - Intrinsic Reward: 5, Ess: False,Imp: 5, Time: 1
\end{verbatim}
\item{2 goals where all tasks are expected to be completed well within the deadline, the goals vary in their value associated for completion}
\begin{verbatim}
 Goal A - Value: 100, Deadline: 20

     SG A1 - Intrinsic Reward: 0, Ess: True,Imp: 90, Time: 3

        Task A11 - Intrinsic Reward: 0, Ess: True,Imp: 30, Time: 1

        Task A12 - Intrinsic Reward: 0, Ess: True,Imp: 30, Time: 1

        Task A13 - Intrinsic Reward: 0, Ess: True,Imp: 30, Time: 1

     SG A2 - Intrinsic Reward: 0, Ess: True,Imp: 10, Time: 3

        Task A21 - Intrinsic Reward: 0, Ess: True,Imp: 3, Time: 1

        Task A22 - Intrinsic Reward: 0, Ess: True,Imp: 3, Time: 1

        Task A23 - Intrinsic Reward: 0, Ess: True,Imp: 4, Time: 1

 Goal B - Value: 200, Deadline: 20

     SG B1 - Intrinsic Reward: 0, Ess: True,Imp: 90, Time: 3

        Task B11 - Intrinsic Reward: 0, Ess: True,Imp: 30, Time: 1

        Task B12 - Intrinsic Reward: 0, Ess: True,Imp: 30, Time: 1

        Task B13 - Intrinsic Reward: 0, Ess: True,Imp: 30, Time: 1

     SG B2 - Intrinsic Reward: 0, Ess: True,Imp: 10, Time: 3

        Task B21 - Intrinsic Reward: 0, Ess: True,Imp: 3, Time: 1

        Task B22 - Intrinsic Reward: 0, Ess: True,Imp: 3, Time: 1

        Task B23 - Intrinsic Reward: 0, Ess: True,Imp: 4, Time: 1
\end{verbatim}
\item{2 goals where only 1 goal is expected to be completed within the given deadline where the 2 goals differ only in their value associated for completion}
\begin{verbatim}
 Goal A - Value: 100, Deadline: 8

     SG A1 - Intrinsic Reward: 0, Ess: True,Imp: 90, Time: 3

        Task A11 - Intrinsic Reward: 0, Ess: True,Imp: 30, Time: 1

        Task A12 - Intrinsic Reward: 0, Ess: True,Imp: 30, Time: 1

        Task A13 - Intrinsic Reward: 0, Ess: True,Imp: 30, Time: 1

     SG A2 - Intrinsic Reward: 0, Ess: True,Imp: 10, Time: 3

        Task A21 - Intrinsic Reward: 0, Ess: True,Imp: 3, Time: 1

        Task A22 - Intrinsic Reward: 0, Ess: True,Imp: 3, Time: 1

        Task A23 - Intrinsic Reward: 0, Ess: True,Imp: 4, Time: 1

 Goal B - Value: 200, Deadline: 8

     SG B1 - Intrinsic Reward: 0, Ess: True,Imp: 90, Time: 3

        Task B11 - Intrinsic Reward: 0, Ess: True,Imp: 30, Time: 1

        Task B12 - Intrinsic Reward: 0, Ess: True,Imp: 30, Time: 1

        Task B13 - Intrinsic Reward: 0, Ess: True,Imp: 30, Time: 1

     SG B2 - Intrinsic Reward: 0, Ess: True,Imp: 10, Time: 3

        Task B21 - Intrinsic Reward: 0, Ess: True,Imp: 3, Time: 1

        Task B22 - Intrinsic Reward: 0, Ess: True,Imp: 3, Time: 1

        Task B23 - Intrinsic Reward: 0, Ess: True,Imp: 4, Time: 1

\item{2 goals where only 1 goal is expected to be completed within the given deadline where the 2 goals differing only in the intrinsic rewards}

 Goal A - Value: 100, Deadline: 8

     SG A1 - Intrinsic Reward: 3, Ess: True,Imp: 90, Time: 3

        Task A11 - Intrinsic Reward: 1, Ess: True,Imp: 30, Time: 1

        Task A12 - Intrinsic Reward: 1, Ess: True,Imp: 30, Time: 1

        Task A13 - Intrinsic Reward: 1, Ess: True,Imp: 30, Time: 1

     SG A2 - Intrinsic Reward: 3, Ess: True,Imp: 10, Time: 3

        Task A21 - Intrinsic Reward: 1, Ess: True,Imp: 3, Time: 1

        Task A22 - Intrinsic Reward: 1, Ess: True,Imp: 3, Time: 1

        Task A23 - Intrinsic Reward: 1, Ess: True,Imp: 4, Time: 1

 Goal B - Value: 100, Deadline: 8

     SG B1 - Intrinsic Reward: 6, Ess: True,Imp: 90, Time: 3

        Task B11 - Intrinsic Reward: 2, Ess: True,Imp: 30, Time: 1

        Task B12 - Intrinsic Reward: 2, Ess: True,Imp: 30, Time: 1

        Task B13 - Intrinsic Reward: 2, Ess: True,Imp: 30, Time: 1

     SG B2 - Intrinsic Reward: 6, Ess: True,Imp: 10, Time: 3

        Task B21 - Intrinsic Reward: 2, Ess: True,Imp: 3, Time: 1

        Task B22 - Intrinsic Reward: 2, Ess: True,Imp: 3, Time: 1

        Task B23 - Intrinsic Reward: 2, Ess: True,Imp: 4, Time: 1
\end{verbatim}
\end{enumerate}